\documentclass{article}
\usepackage{tikz}
\usepackage{amsmath,amssymb,amsthm,bm,enumerate,color,geometry,mathrsfs,ulem}
\usepackage{graphicx}
\graphicspath{{./Figure}}
\usepackage{ascmac}
\usepackage[titletoc,title]{appendix}
\usepackage{algorithm, algorithmic}

\newcommand{\keywords}[1]{\textbf{Key words:} #1}

\def\disp{\displaystyle}

\def\CA{\mathcal{A}}

\def\rnum#1{\expandafter{\romannumeral #1}}
\def\Rnum#1{\uppercase\expandafter{\romannumeral #1}}

\title{On Choice of Hyper-parameter in Extreme Value Theory based on Machine Learning Techniques}

\author{Chikara Nakamura \thanks{Mathematical Science Research Laboratory, Nikon Corporation, 244-0843, Japan. \ Email: chikara.nakamura@nikon.com}}
\date{July 13th, 2021}

\begin{document}

\newtheorem{Def}{Definition}[section]
\newtheorem{Prop}[Def]{Proposition}
\newtheorem{Thm}[Def]{Theorem}
\newtheorem{Ass}[Def]{Assumption}
\newtheorem{Lem}[Def]{Lemma}
\newtheorem{Rem}[Def]{Remark}
\newtheorem{Example}[Def]{Example}
\newtheorem{Cor}[Def]{Corollary}
\newtheorem{Notation}[Def]{Notation}
\newtheorem{Algorithm}[Def]{Algorithm}
\newtheorem{Condition}[Def]{Condition}
\newtheorem*{Rem*}{Remark}
\makeatletter
\@addtoreset{equation}{section}
\def\theequation{\thesection.\arabic{equation}}
\makeatother
\maketitle
%\tableofcontents

%% Abstract
\begin{abstract}
Extreme value theory (EVT) is a statistical tool for analysis of extreme events.
It has a strong theoretical background, however, we need to choose hyper-parameters
 to apply EVT.
In recent studies of machine learning, techniques of choosing hyper-parameters have been well-studied.
In this paper, we propose a new method of choosing hyper-parameters in EVT based on machine learning techniques.
We also experiment our method to real-world data and show good usability of our method.
\end{abstract}

\begin{flushleft}
  \\
% \msc{60J10, 60J20.}  \\
% \reviseddate{2021/07/04}
\end{flushleft}

\section{Introduction}  \label{Sec:Introduction}
Using data in real-world applications has been attracting many people's attentions.
Anomaly events, which deviate from normal patterns, naturally arise in real life applications such as network intrusion detection, insurance and finance.
Recently, analysis of such events has been getting more important.

\medskip
% Overview of EVT (theory and application)
In mathematical statistics, EVT is a powerful tool for analysis of anomaly events.
For random variables $\{ X_t \}_t$ which independently and identically distributed with a random variable $X$, EVT gives properties of statistics such as the maximum $\disp \max X_t$ and the excess over threshold $X - u \mid X>u$ (see Section \ref{Sec:RevEVT} for the details). EVT has a long history of its research, and it  has a strong theoretical background (see \cite{CBTD01, EKM13} and references therein).
In addition, as we will see in Section \ref{Sec:RelatedWork}, EVT has been used for real world analysis such as finance, epidemics and meteorology.

\medskip
% problem of choice of hyper-parameter
In this paper, we focus on the peak over threshold (POT) method, which
is one of the methods of extreme value analysis.
POT method fits data exceeding over a threshold by the generalized Pareto distribution (GP).

To apply POT method, we need to choose a threshold.
Choice of hyper-parameters is a difficult task since we need to take into accounts
the factors such as number of sample data, fitness of the theory, accuracy and stability of estimates.
In EVT, graphical diagnostics are commonly used for choice of hyper-parameters.
However, these methods are subjective and do NOT uniquely determine the values of hyper-parameters.
In author's opinion, these methods have a lot of room of improvement from the viewpoint of practice.
In fact, a lot of methods of choosing hyper-parameters have been proposed (see
Section \ref{Sec:RelatedWork}).

\medskip
% What we consider in this paper
Machine learning, which is a branch of artificial intelligence, has lately attracted increasing attention due to its wide applications in many areas.
In the context of machine learning, choice of threshold can be regarded as a problem of hyper-parameters.
Choice of hyper-parameters is one of the strong assets of machine learning.
In this paper, we propose a new method for choice of threshold in EVT based on machine learning.

\medskip
% why machine learning
As we mentioned above, there are a lot of methods of choice of hyper-parameter in EVT.
These criteria are based on theory of statistics and explicit formula derived from EVT.
In contrast, machine learning enables us to design evaluation indicator more flexibly.
As a result, we can easily incorporate factors that we want to take into account for criteria of choice of hyper-parameters, and we can improve methods of choice of hyper-parameter in more practical way.
In fact, after reviewing the related works (see Section \ref{Sec:RelatedWork}) and
overviewing EVT (see Section \ref{Sec:RevEVT}),
we introduce an evaluation indicator which balances the fitness of EVT and accuracy of estimate.
We adopt the value that the indicator attains its minimum as the hyper-parameter.
The indicator doesn't have an analytic representation, however, we can optimize the indicator based on machine learning techniques.
Our method automatically and uniquely determine the value of hyper-parameter, which reduce the difficulty of applying EVT.

\medskip
% real-world data
We demonstrate our method for both synthetic and real-world data.
We show the validity of our method by probability plot, quantile plot, return level plot and density plot (see Section \ref{Sec:Check} for the details).

\medskip
% future work
%We can find a lot of analysis and applications of EVT.
%However, to the best of the author's knowledge, there is few research of enhancing the usability of EVT itself.
%In the future, we will be able to feedback our method to the applications of EVT such as \cite{SFTL17, YXY18}.

% features of this approach
The features of this paper are the following:
\begin{enumerate}  \renewcommand{\labelenumi}{(\arabic{enumi})}
    \item  Machine learning is one of the fields of data analysis, and there are a lot of research which incorporate theory of statistics and mathematics into research of machine learning.
Our method is reverse in the sense that we incorporate machine learning techniques into the theory and improve its usability.
In the future, we will be able to feedback our method to the applications of EVT such as \cite{SFTL17, YXY18}.

   \item  We discuss a method of choice of hyper-parameter in EVT.
   As we will see in Section \ref{Sec:RelatedWork}, there are a lot of research of choice of hyper-parameter in EVT.
   However, there is few research based on machine learning.

\end{enumerate}

\medskip
We organize this paper as follows:
After reviewing the related works,
we provide an overview of EVT in section \ref{Sec:RevEVT}.
In section \ref{Sec:Method}, we describe our method of the choice of hyper-parameter.
In section  \ref{Sec:NumStudy}, we demonstrate our method with synthetic and  real-world data.
Finally in section \ref{Sec:Discuss}, we discuss the results of our method and future works.

\subsection{Related work}  \label{Sec:RelatedWork}

The study of extreme value theory was initiated after the
 flood of the North Sea from the night of January 31 to February 1 in 1953,
  leading to the death of more than 1800 people in the Netherlands alone.
After that, the theory of statistics of rare events was extensively studied.
Extreme value theory has been used in various applications such as finance \cite{Marco14},  appearance of epidemics \cite{CLZP15} and  climatology \cite{KZZH07}.
 In \cite{CBTD01}, we can find a lot of examples of real-world applications of EVT.
Recently, extreme value theory is also applied to anomaly detection
 \cite{SFTL17, YXY18, EE20}, which is extensively studied in the field of machine learning.

\medskip
In POT method, we fit observation
$X_1, \ldots, X_n$ over a threshold $u$ by GPD.
We need to choose a hyper-parameter $u$ to apply POT method.
Many procedures for hyper-parameter selection have been proposed.
We briefly review these methods in this section.
For more details, see the review papers \cite{SM12, CG16, LMPD16}

%(see \cite{SM12, LMPD16} for examples),
%from intuitive methods to more sophisticated ones.

\medskip
Rules of thumbs include selecting the top 10\% of the data \cite{DuMouchel83} and top  square root \cite{FDP03}.
Such methods are often employed even though they are inappropriate from the viewpoint of theory.
Graphical diagnostics are also popular for selection of hyper-parameter \cite{CBTD01}.
This category is based on the fact that the mean of excesses $E[X-u \mid X>u]$ linearly depends on $u$.
As mentioned in \cite{SM12, BYZ18}, a key drawback with these approaches is that such methods are subjective and and are difficult to apply consistently.

\medskip
A different category of methods is based on goodness-of-fit test \cite{DS90,CS01,BYZ18}.
The methods in this category select the hyper-parameter as the lowest value above which the EVT provides adequate fit to the excess over the value.
Hill's estimator \cite{Hill75} is based on tail index of the Pareto
type distribution.
In \cite{CS01}, $W^2$ C\'{r}amer-von Mises and $A^2$ Anderson-Darling statistics are used for
goodness-of fit test.

\medskip
We can give other categories for choice of hyper-parameters,
 such as mixture models, bootstrap and Bayesian inference.
To the author's best knowledge, there are few research from the viewpoint of machine learning.
In the following sections, we propose a new method based on machine learning techniques.

\section{Brief review of EVT} \label{Sec:RevEVT}
Extreme value theory focuses on statistics of rare events.
In this section, we provide a brief overview of EVT which is necessary for this paper.
The main interest of this paper is POT method, however,
  we provide another popular approach, the block maxima (BM) model, for readers' convenience.
For more details, see \cite{CBTD01, EKM13} and references therein.

\subsection{The extreme value distribution}
Let $X_t$ ($t=1,2, \ldots$) be an i.i.d. random variables with distribution function $F$.
In addition, define $\displaystyle Z_t := \max_{1 \le s \le t} X_s$.
The Fisher–Tippett–Gnedenko theorem \cite{FT28, Gnedenko43} states that there exist sequences $\{ a_t \}_{t=1}^{\infty}$ and $\{ b_t \}_{t=1}^{\infty}$ and a non-degenerated random variable $Z$ such that
\begin{align*}
	P \left( \frac{Z_t - b_t}{a_t} \le x \right) \to P(Z \le x) =: G(x) \,.
\end{align*}
The limit distribution $G(x)$ is called the generalised extreme value distribution and is given by
\begin{align} \label{eq:distGEV}
    G(x) =
    \exp \left\{-\left[1+\xi\left( \frac{x-\mu}{\sigma}\right)\right]^{-1 / \xi} \right\} \,, \qquad  \mu \in \mathbb{R}, \sigma>0, \xi \in \mathbb{R} \,.
\end{align}

\medskip
There is another important distribution, which is called the generalized Pareto (GP) distribution, in EVT.
The statistics of exceeding data over a threshold $u$ is of interest.
The Pickands–Balkema–de Haan theorem \cite{BH74, Pickands75} states that
\begin{align*}
\bar{F}_{u}(x)=\mathbb{P}(X-u>x \mid X>u) \sim \left(1+\frac{\gamma x}{\sigma(t)}\right)^{-\frac{1}{\gamma}} \qquad \text{for sufficient large $u$.}
\end{align*}
The limit distribution above is called the GP distribution
and is given by
\begin{align*}
    H(y) = H_{\xi} \left( \frac{y}{\sigma} \right) =
    \begin{cases}
        1 - \left( 1 + \xi \frac{y}{\sigma} \right)_+^{-1/\xi}  \,, & \xi \neq 0 \,,  \\
        1 - \exp \left ( - \frac{y}{\sigma} \right) \,,  &  \xi = 0 \,,
    \end{cases}
\end{align*}
So, the probability law of $X - u \mid X>u$ can be approximated by (GP) for sufficiently large $u$.

\subsection{Block Maxima and Peaks Over Threshold methods}
In analysis using extreme value theory, the following two models are mainly used:
BM and POT methods.
In this section, we briefly overview the two approaches.

\begin{flushleft} {\bf The BM method} \end{flushleft}
The BM approach consists of dividing the observation period into $k$ blocks of each length $m$ and
 restricts attention to the maximum observation in each block.
More specifically, we define
\begin{align*}
    Y_{i}=\max _{(i-1) m<j \leq i m} X_{j}
\end{align*}
If $m$ is sufficiently large, then the distribution of $Y_i, (i=1,2,\ldots)$ can be approximated by the GEV distribution.
We can estimate the parameters $\mu, \sigma, \xi$ in the GEV distribution by the maximal likelihood estimation.
The log-likelihood function is given by
\begin{align*}
 \ell(\mu, \sigma, \xi)=
    -n \log \sigma-(1 / \xi+1) \sum_{i=1}^{n} \log \left[1+\xi\left(\frac{y_{i}-\mu}{\sigma}\right)\right]
     -\sum_{i=1}^{n} \left[ 1+\xi\left(\frac{y_{i}-\mu}{\sigma}\right)\right]^{-1 / \xi}
\end{align*}
provided  $\xi \neq 0$ and  $\disp 1+\xi\left(z_{i}-\mu\right) / \sigma>0$ $(i=1,2, \ldots, n)$, and
\begin{align*}
     \ell(\mu, \sigma, 0)
         = -n \log \sigma-\sum_{i=1}^{n}\left(\frac{y_{i}-\mu}{\sigma}\right)-\sum_{i=1}^{n} \exp \left[-\left(\frac{y_{i}-\mu}{\sigma}\right)\right]  \,.
\end{align*}
The maximal likelihood estimators $\hat{\mu}, \hat{\sigma}, \hat{\xi}$ are given by
\begin{align*}
    \hat{\mu}, \hat{\sigma}, \hat{\xi} = \arg \max_{\mu, \xi, \sigma} \ell (\mu, \sigma, \xi) \,.
\end{align*}

\begin{flushleft} {\bf The POT approach} \end{flushleft}

In the POT method, the excess data over a threshold $u$ is of interest.
Let denote by $\{ y_1, y_2, \ldots, y_m \}$ the set of excesses over the threshold $u$.
If $u$ is sufficiently large, then the distribution of $\{ y_1, y_2, \ldots, y_m \}$ can be approximated by the GP distribution.

\medskip
The log-likelihood of GP model is given by
\begin{align*}
    \ell(\sigma, \xi)=
    \begin{cases} \displaystyle
        -n \log \sigma-(1 / \xi+1) \sum_{i=1}^{n} \log \left(1+\xi y_{i} / \sigma\right)
            &  \text{if $\xi \neq 0$ and } 1+\xi y_{i} / \sigma>0, \quad i=1,2, \ldots, n \,, \\
        \displaystyle -n \log \sigma-\frac{1}{\sigma} \sum_{i=1}^{n} y_{i}
            &  \text{if $\xi = 0$.}
    \end{cases}
\end{align*}
The maximal likelihood estimators $\hat{\sigma}, \hat{\xi}$ are given by
\begin{align*}
    \widehat{\xi}, \widehat{\sigma} = \arg \max_{\xi, \sigma} \ell (\xi, \sigma) \,.
\end{align*}

\subsection{Model checking of POT method} \label{Sec:Check}
\begin{figure}[htbp]
  \begin{center}
    \begin{tabular}{c}
      \begin{minipage}{0.5\hsize}
        \begin{center}
          \includegraphics[clip, width=6cm]{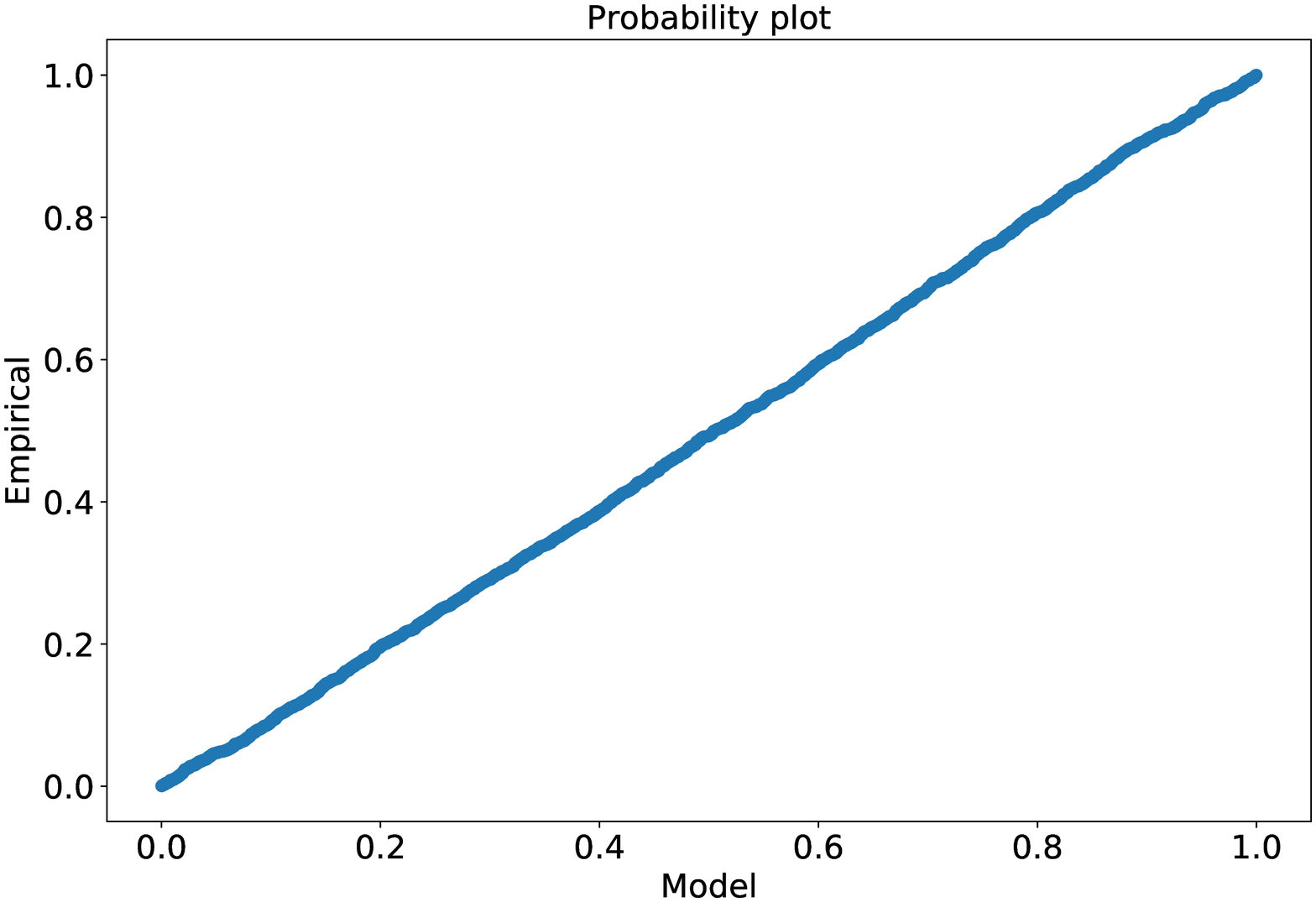}
        \end{center}
      \end{minipage}

      \begin{minipage}{0.5\hsize}
        \begin{center}
          \includegraphics[clip, width=6cm]{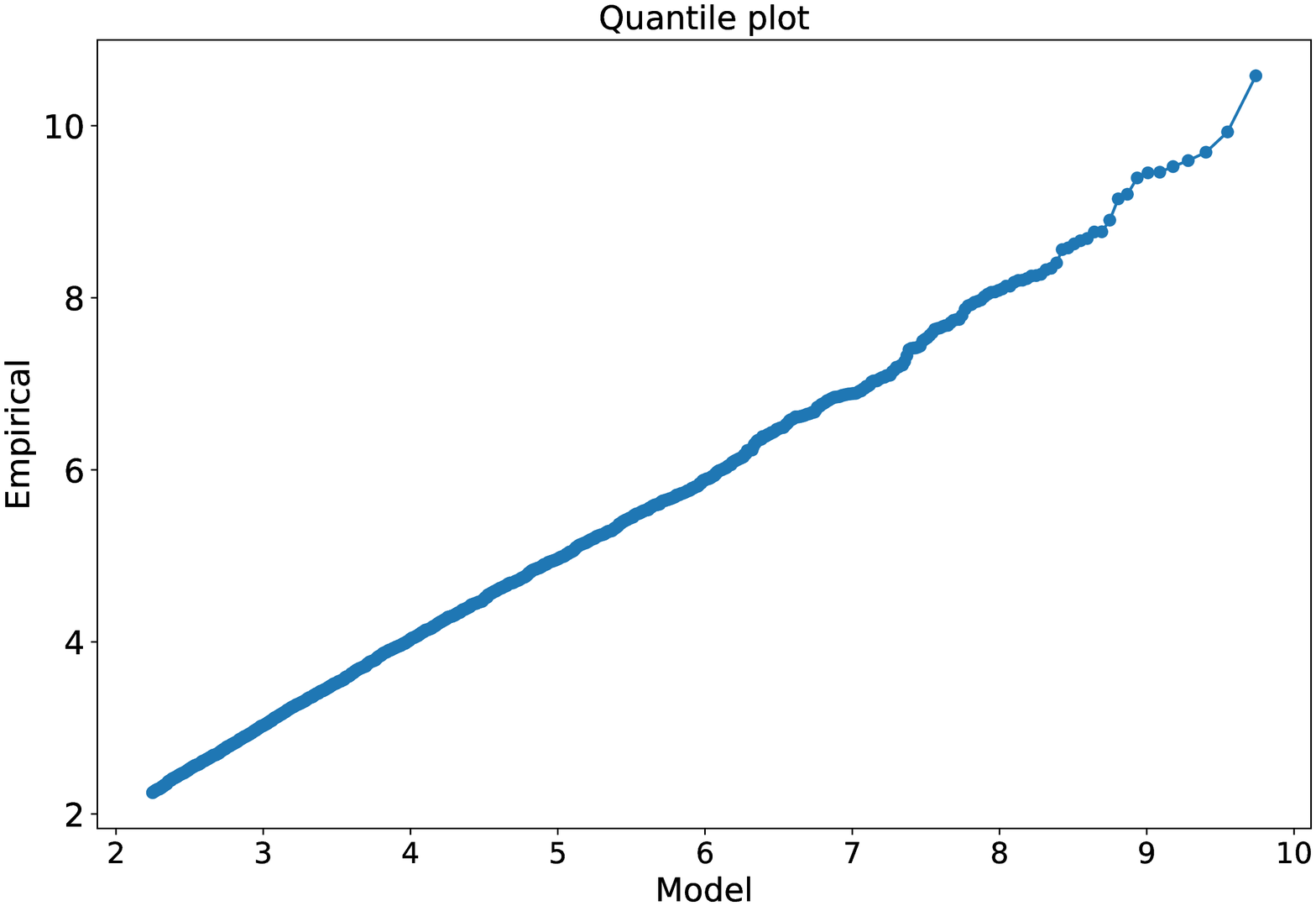}
        \end{center}
      \end{minipage}\\
      \begin{minipage}{0.5\hsize}
        \begin{center}
          \includegraphics[clip, width=6cm]{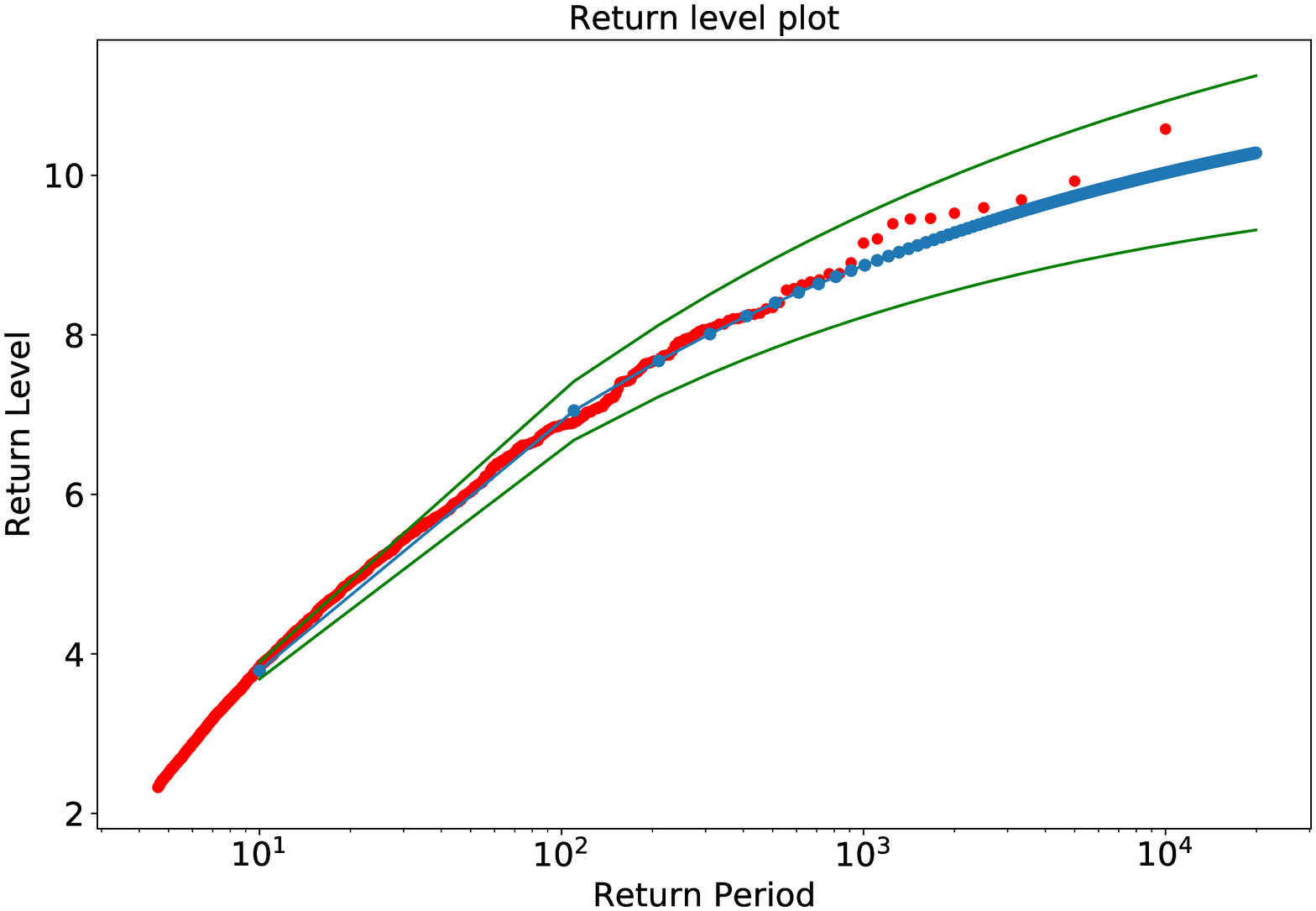}
        \end{center}
      \end{minipage}

      \begin{minipage}{0.5\hsize}
        \begin{center}
          \includegraphics[clip, width=6cm]{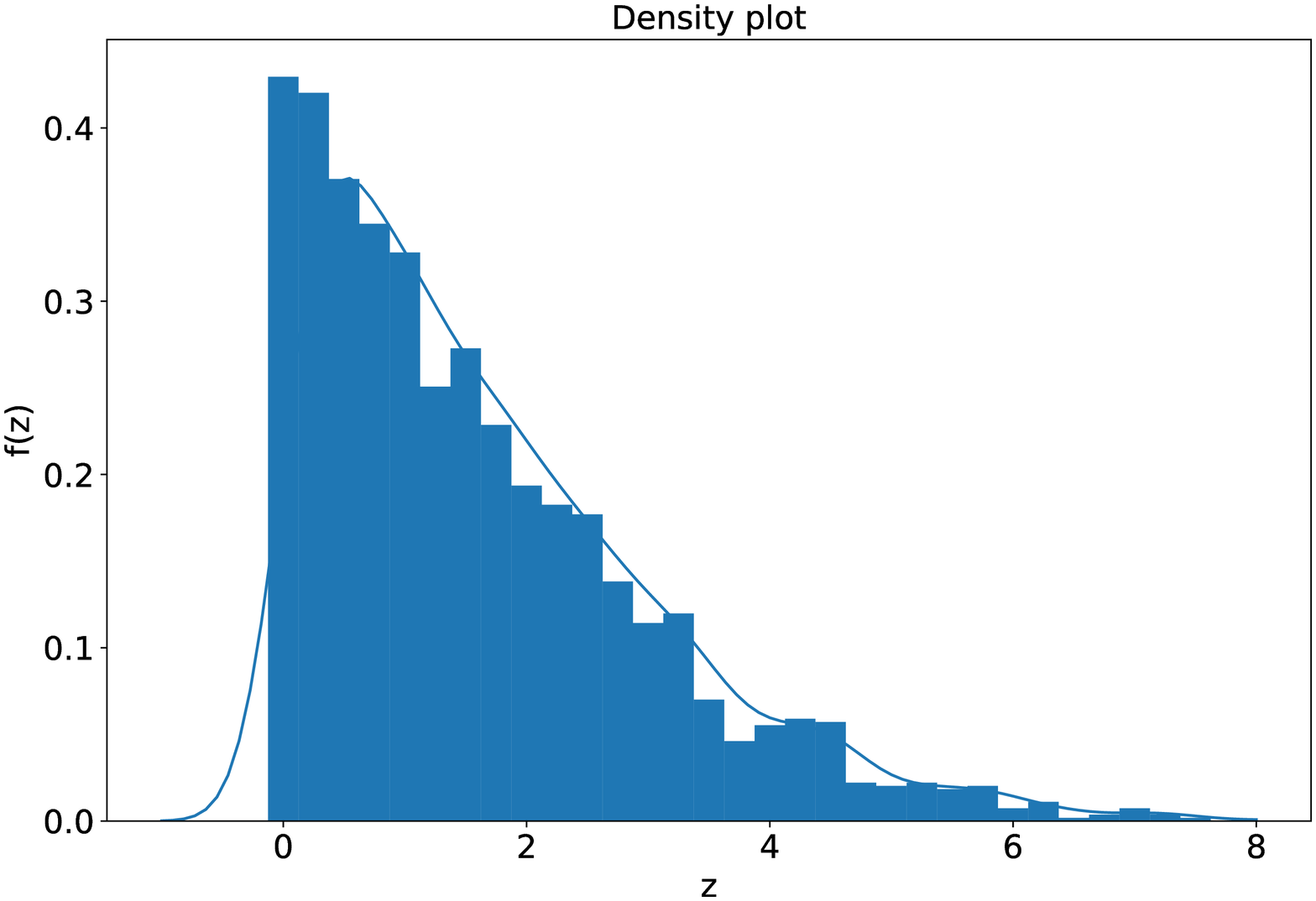}
        \end{center}
      \end{minipage}\\

    \end{tabular}
    \caption{Illustrations of probability plot, quantile plot, return level plot and density plot for the $u$ selected above.}
    \label{Fig:Check_Gauss}
  \end{center}
\end{figure}

From now on, we focus only on POT method.
To evaluate the fitness, the following four plots are often used: probability plots, quantile plots, return level plots and density plots (see Figures \ref{Fig:Check_Gauss} and \ref{Fig:Check_Gam1}).

\medskip
Let $ y_{(1)} \le \cdots  \le y_{(k)}$ be the excesses over a threshold $u$ and $\widehat{\sigma}, \widehat{\xi}$ be the parameters of GP distribution estimated by the excesses.
The probability plot consists of the pairs
\begin{align*}
   \left\{\left(i /(k+1), \hat{H}\left(y_{(i)}\right)\right) ; i=1, \ldots, k\right\}\,,
   \qquad \text{where} \qquad  \hat{H}(y)=1-\left(1+\frac{\hat{\xi} y}{\hat{\sigma}}\right)^{-1 / \hat{\xi}} \,.
\end{align*}
Similarly, the quantile plot consists of the pairs
\begin{align*}
\left\{\left(\hat{H}^{-1}(i /(k+1)), y_{(i)}\right), i=1, \ldots, k\right\} \,, \qquad
\text{where} \qquad \hat{H}^{-1}(y)=u+\frac{\hat{\sigma}}{\hat{\xi}}\left[y^{-\hat{\xi}}-1\right]
\end{align*}
Both the probability and quantile plots should consist of points that are approximately linear
  if the POT method fits properly.

\medskip
A return level plot consists of the pair $\{(m, \widehat{x}_m) \}$, where
\begin{align*}
    \widehat{x}_m =u+\frac{\hat{\sigma}}{\hat{\xi}}\left[\left(m \hat{\zeta}_{u}\right)^{\hat{\xi}}-1\right]
\end{align*}
and $\widehat{\zeta}_u$ is the ratio of the number of excesses over the threshold $u$ to the total number of samples.

\medskip
A density plot compares the density function of GP distribution and histogram of excesses over a threshold.

\subsection{Choice of threshold} \label{Sec:Choice}
To apply POT method, we need to choose a threshold $u$, which is a hyper-parameter of POT model.
In EVT, there are two main criteria for the choice of threshold $u$.

\begin{flushleft} {\bf The mean excess plot} \end{flushleft}
Let $x_{(1)}, x_{(2)}, \ldots, x_{(n_u)}$ be sample data over a threshold $u$.
The mean excess plot consists of the pair
\begin{align*}
    \left\{\left(u, \frac{1}{n_{u}} \sum_{i=1}^{n_{u}}\left(x_{(i)}-u\right)\right): u<x_{\max }\right\} \,,
\end{align*}
The mean excess plot should be approximately linear in $u$ in a range where the generalized Pareto distribution approximate the excess distribution.
We choose the value of threshold as the highest of $u$ which the mean excess plot is linear.

\begin{flushleft} {\bf Stability of $\widehat{\xi}$ and $\widehat{\sigma}$} \end{flushleft}
Another criterion for the choice of threshold is stability of estimate.
Denoting by $\widehat{\xi}$ and $\widehat{\sigma}$ the MLE of the parameters $\xi$ and $\sigma$ of GP distribution respectively,
  both $\widehat{\xi}$ and $\widehat{\sigma}_{\ast} = \widehat{\sigma} - u \widehat{\xi}$ against $u$ should be constants.
So, plotting both $\widehat{\xi}$ and $\widehat{\sigma}_{\ast}$ together with confidence intervals for each of these quantities,
 and selecting the threshold $u$ as the estimates remain near-constant.

\begin{flushleft} {\bf Limitation of the above approaches} \end{flushleft}

As we discussed in Section \ref{Sec:Introduction},
the above graphical diagnostics are subjective and do NOT uniquely determine the value of hyper-parameter.
In the the following sections, we describe our method of choice of the threshold $u$.

\section{Methodologies} \label{Sec:Method}
In this section, we describe our new method for the choice of threshold $u$.

\subsection{Kernel density estimation (KDE)}  \label{Sec:KDE}
KDE is a non-parametric method of estimating the underlying probability density function of a dataset.
Let $\{ x_i \}_{i=1}^N$ be a data driven by an unknown probability density function $p(x)$.
Then, KDE at a point $y$ is given by
\begin{align*}
    \rho_{K}(y)=\sum_{i=1}^{N} K\left(y-x_{i} ; h\right)
\end{align*}
where $K$ is a non-negative function which satisfies the following:
\begin{align*}
    \int K(x) \mathrm{d} x=1, \quad \int x K(x) \mathrm{d} x=0, \quad \int x^{2} K(x)>0
\end{align*}
The function $K$ above is called the kernel function.
We adopt the Gaussian kernel, which is given by
    \begin{align*}
        K(x)=\frac{1}{\sqrt{2 \pi}} \exp \left(-\frac{x^{2}}{2}\right) \,.
    \end{align*}

\medskip
In this paper, we use KDE to estimate GP distribution, which has a discontinuous point at $x=0$.
We employ the following modified KDE to enhance the approximation of KDE:
\begin{align*}
    \widetilde{\rho}_{K}(y) := \frac{\rho_{K}(y - \epsilon)}{\int_{y > \epsilon} \rho_{K}(y - \epsilon) dy} \,,
\end{align*}
where $\epsilon > 0$ is a sufficiently small positive constant.

\subsection{Score}
Before moving on the details, we give an intuition behind our method.

\medskip
Fig \ref{Fig:Idea} are illustrations of time series generated by i.i.d. Gaussian distribution with different values of threshold.
We estimate the probability distribution of excesses over threshold by MLE and KDE.

\medskip
In the leftmost case, the threshold $u$ is too large, and the number of excesses is NOT sufficient to estimate the probability law.
Hence, the results of MLE and KDE are different from one another.
In contrast, in the middle case,  the number of excesses is sufficient for both MLE and KDE, and hence
  the results of MLE and KDE are similar to each other.
In the rightmost case,  the threshold $u$ is too small, and the excesses over the threhold do NOT fit to the GP distribution.
As a result, the distribution estimated by MLE and KDE are different from one another again.

\begin{figure}[htbp] \label{Fig:Idea}
  \begin{center}
    \begin{tabular}{c}
      \begin{minipage}{0.33\hsize}
        \begin{center}
          \includegraphics[clip, width=6cm]{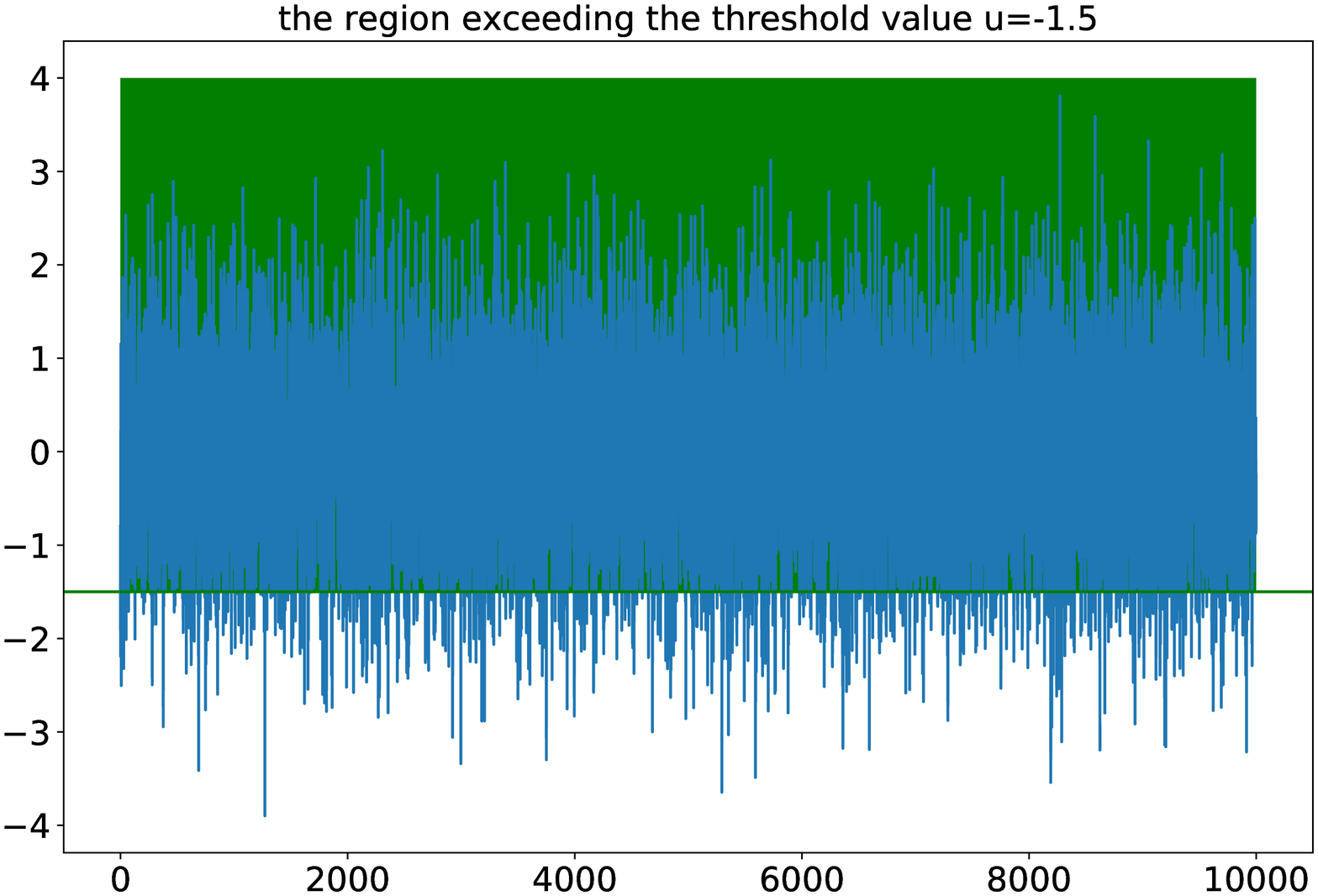}
        \end{center}
      \end{minipage}

      \begin{minipage}{0.33\hsize}
        \begin{center}
          \includegraphics[clip, width=6cm]{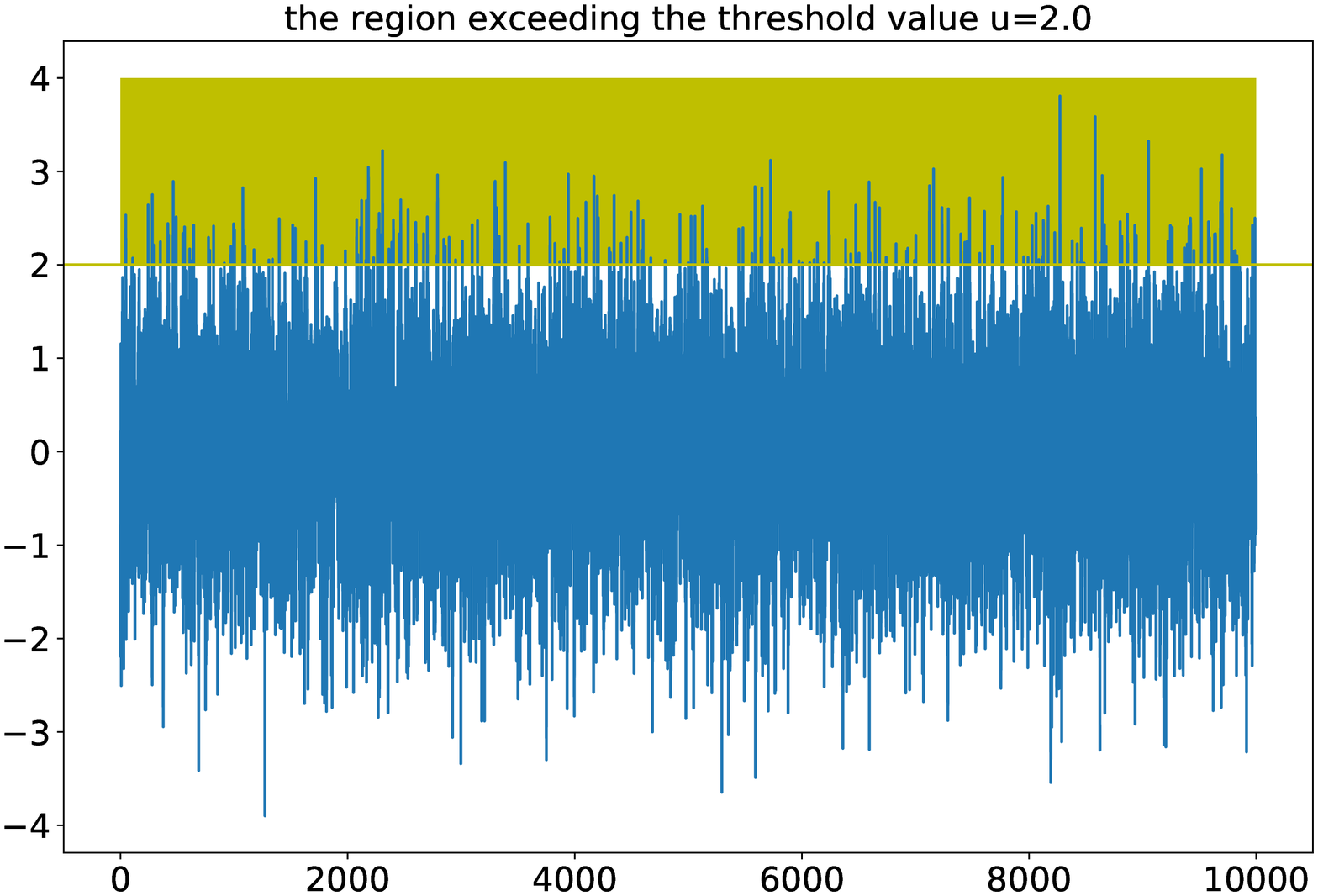}
        \end{center}
      \end{minipage}

      \begin{minipage}{0.33\hsize}
        \begin{center}
          \includegraphics[clip,width=6cm]{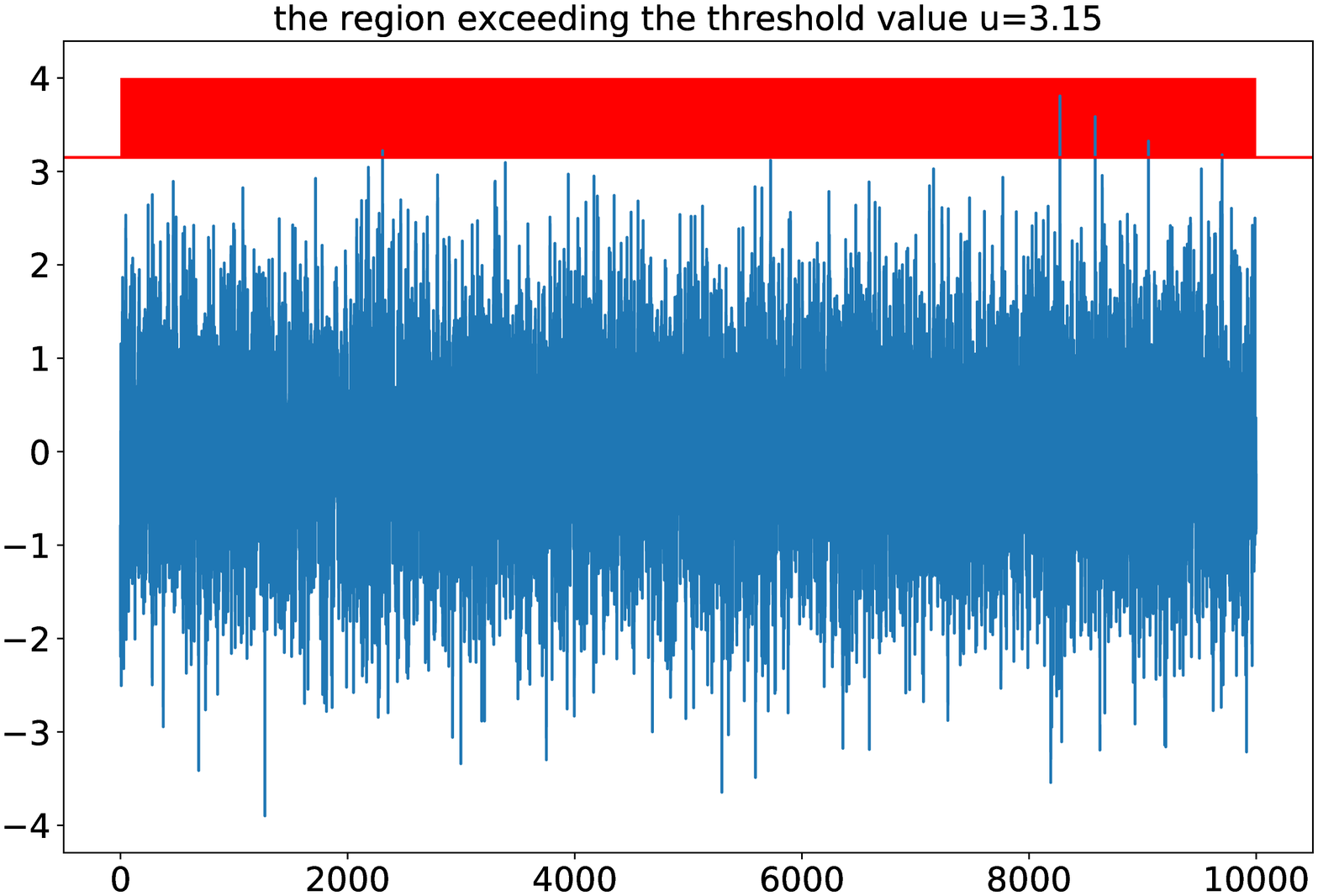}
        \end{center}
      \end{minipage}

    \end{tabular}
    \caption{Illustrations of choice of thresholds}
    \label{Fig:Idea}
  \end{center}
\end{figure}

\begin{figure}[htbp]
  \begin{center}
    \begin{tabular}{c}

      \begin{minipage}{0.33\hsize}
        \begin{center}
          \includegraphics[clip, width=6cm]{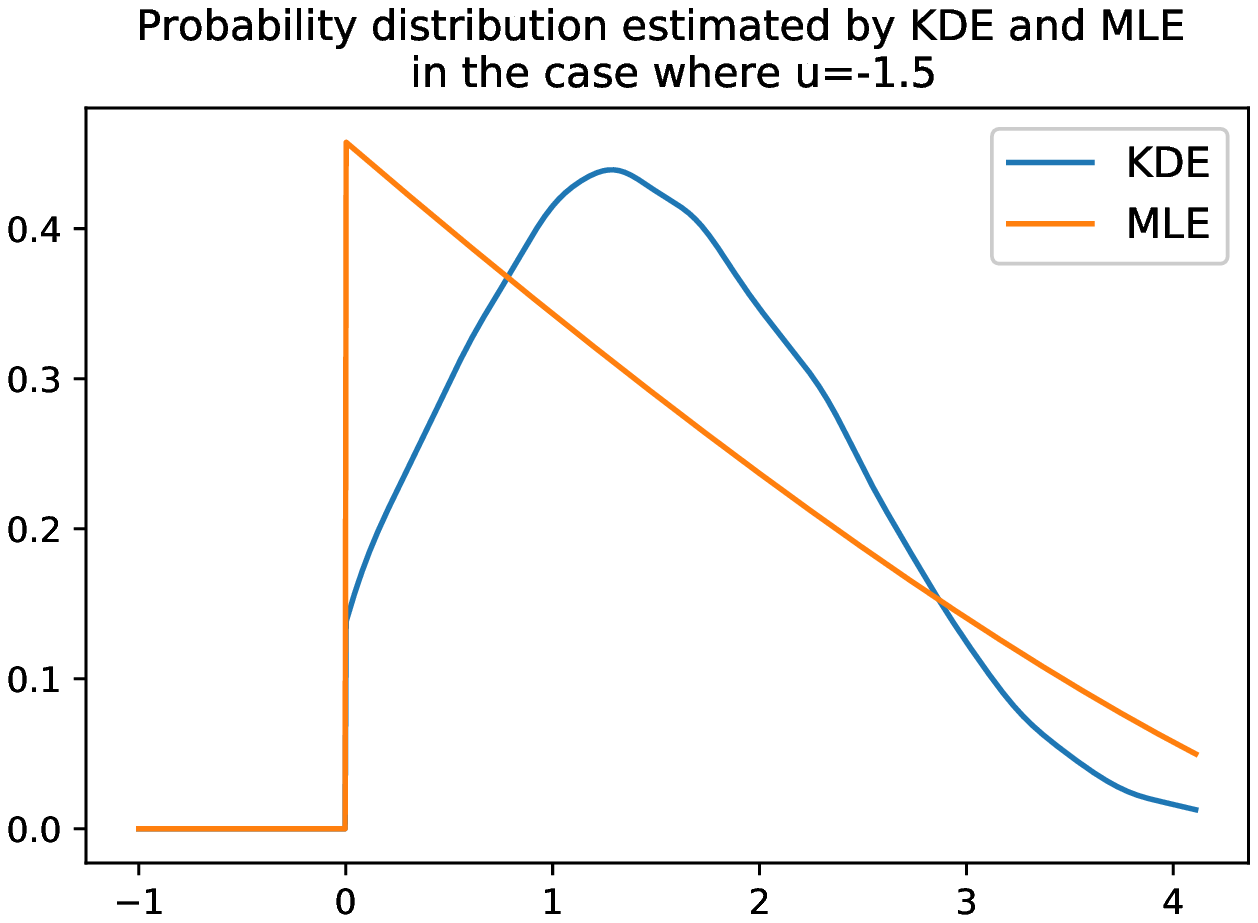}
        \end{center}
      \end{minipage}

      \begin{minipage}{0.33\hsize}
        \begin{center}
          \includegraphics[clip, width=6cm]{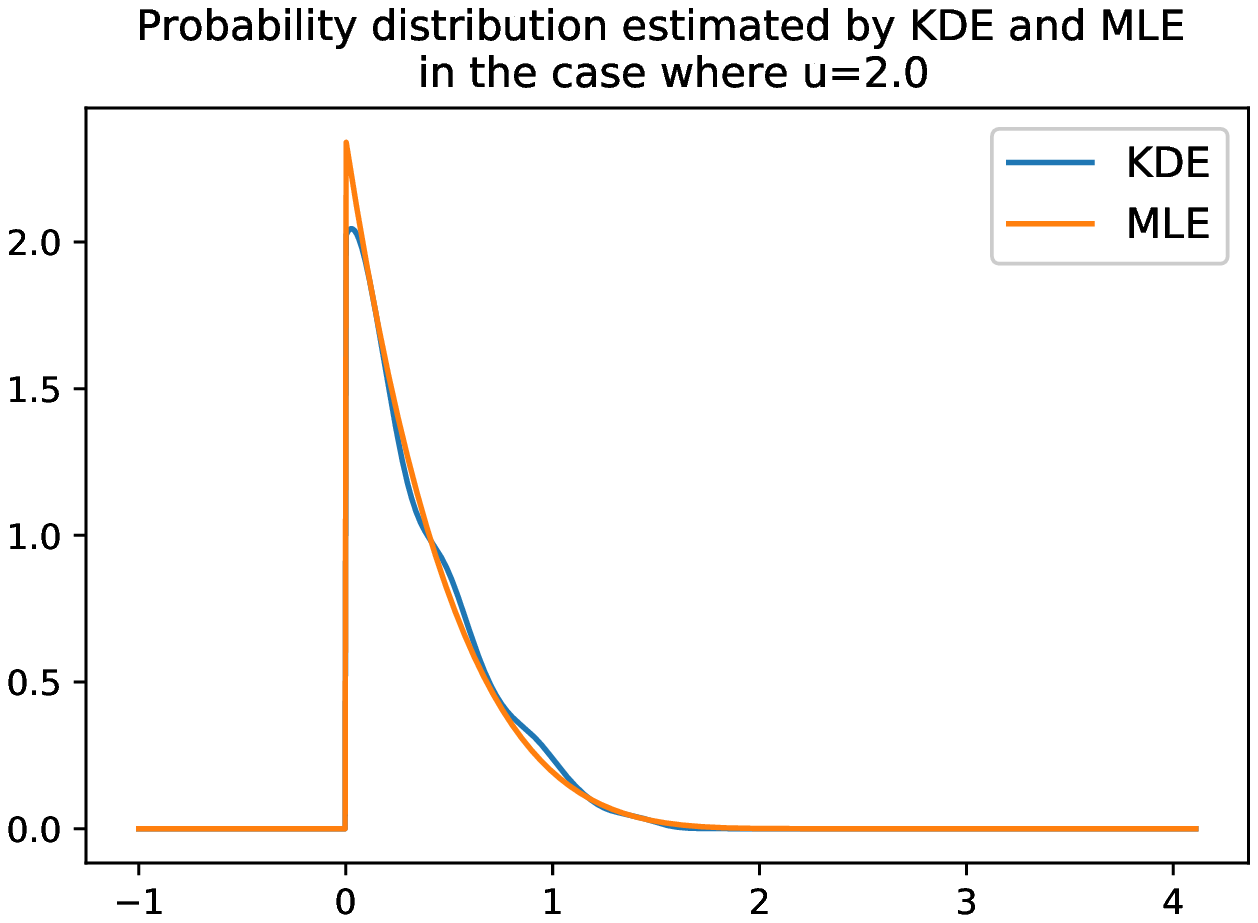}
        \end{center}
      \end{minipage}

      \begin{minipage}{0.33\hsize}
        \begin{center}
          \includegraphics[clip, width=6cm]{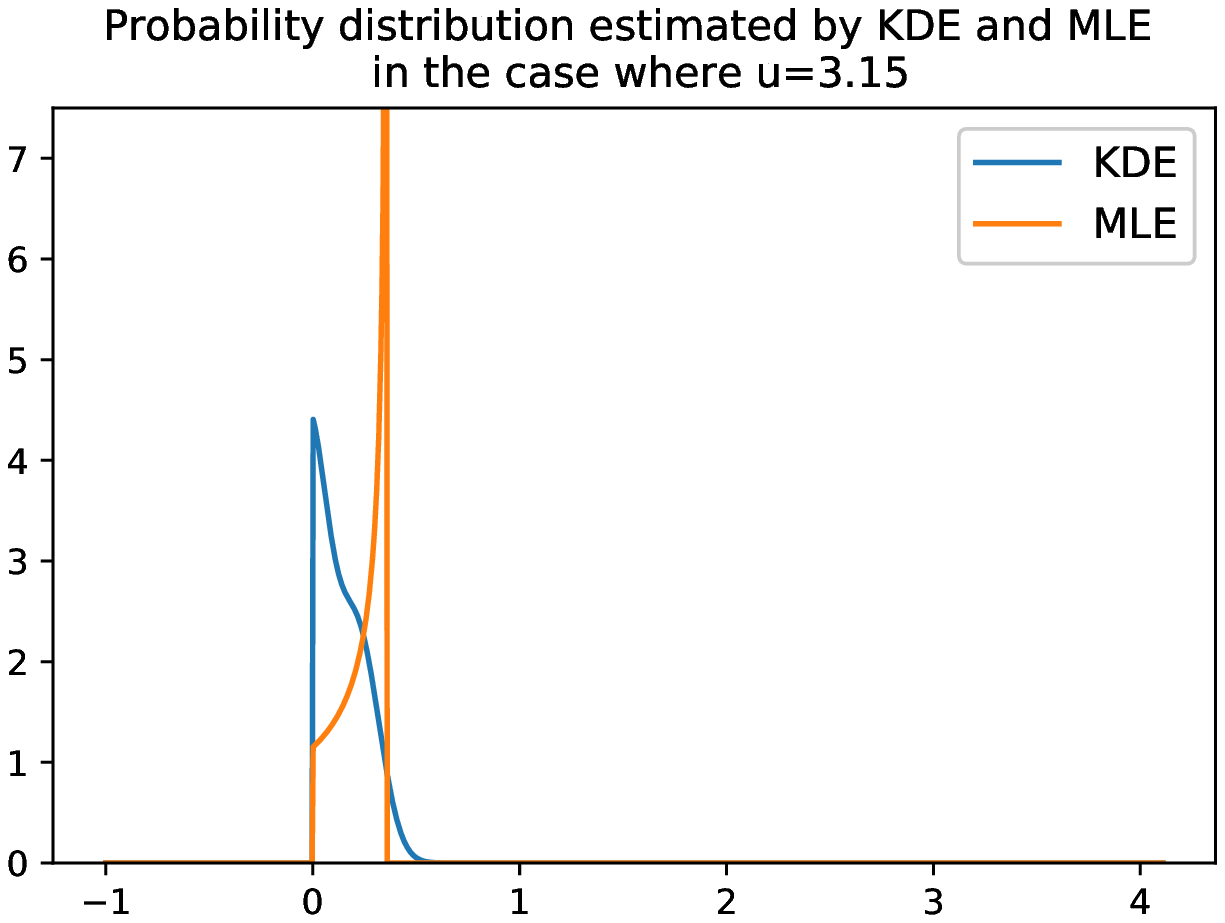}
        \end{center}
      \end{minipage}

    \end{tabular}
    \caption{Comparisons of KDE and MLE with different threshold $u$.}
    \label{Fig:Compare}
  \end{center}
\end{figure}

\medskip
From the above observation, the difference of the probability distribution estimated by MLE and KDE would correspond to the fitness of the threshold.
Based on this idea, we introduce a new score
\begin{align}  \label{eq:Score}
    \operatorname{Score} (u) = \operatorname{Score} (u ;  \{ x_i \}_{i=1}^N ) = C \| \widehat{p}_{kde} (\cdot ) - \widehat{p}_{mle} (\cdot ) \|_1 \,,
\end{align}
denoting by $C$ a positive constant, by $\widehat{p}_{mle}$ and $\widehat{p}_{kde}$ the probability density function estimated by MLE and KDE respectively,
and by $\| \cdot \|_1$ $L^1$ norm.
The score defined in \eqref{eq:Score} measures the difference between $\widehat{p}_{mle}$ and $\widehat{p}_{kde}$,
and we would like to choose the threshold in the way that  $\operatorname{Score} (u)$ is as small as  possible.
We optimize $\operatorname{Score} (u)$ by Bayesian optimization, which is one of the popular method in machine learning.

\subsection{Bayesian optimization (BO)}
BO is one of the machine-learning-based optimization methods to solve optimization problem.
We adopt the value which $\operatorname{Score} (u)$ attains minimum
\begin{align}   \label{eq:BOScore}
    \min_{u \in \CA} \operatorname{Score} (u) \,.
\end{align}

We determine the search range $\CA$ in \eqref{eq:BOScore} with the methods in Section \ref{Sec:Choice},
namely, we use the two criteria, linearity of mean excess plots and stability of $\widehat{\xi}$ and $\widehat{\sigma}$, to determine the domain $\CA$.

\medskip
The reasons of employing BO are listed below:
\begin{enumerate}  \renewcommand{\labelenumi}{(\arabic{enumi})}
    \item By definition, we cannot give an analytic expression of the  $\operatorname{Score}$.

    \item By (1),  we observe only the value of $\operatorname{Score}$ and no first- or second-order derivatives, which are necessary for conventional optimization algorithms.

    \item The objective function $\operatorname{Score}$ is continuous.

    \item   Evaluating  objective function $\operatorname{Score}$  is computation-cost-consuming, since we need to solve an optimization problem to compute $\widehat{p}_{mle}$ for each step.
\end{enumerate}
For the above reasons, BO is a better method than other optimization methods.

\subsection{Summary of our method}
Now, we summarize our method in Algorithm 1.

\begin{algorithm} \label{Alg:OurMethod}
 \caption{Choice of threshold $u$ of POT approach}
 \begin{algorithmic}[1]
 \renewcommand{\algorithmicrequire}{\textbf{Input:}}
 \renewcommand{\algorithmicensure}{\textbf{Output:}}
 \REQUIRE A time series $\{ x_t \}_{t=1}^T$.
 \ENSURE  A threshold $u$.
% \\ \textit{Initialisation} :
  \STATE Set a search range of threshold $u$ by the excess plot and stability of $\widehat{\xi}$ and $\widehat{\sigma}_{\ast}$.

  \STATE Choose initial points $\{ u_i \}_{i=1}^K$ and evaluate $\operatorname{Score}$.
% \\ \textit{LOOP Process}
  \FOR {$j = 1$ to $L$}
  \STATE Apply GPR for $\{ (u_{\ell}, \operatorname{Score} (u_{\ell}) \}_{\ell=1}^{K+j-1}$.
  \STATE Determine the next search point $u_j$ based on EI.
  \STATE Compute $\operatorname{Score}$ at $u_j$.
  \ENDFOR
 \RETURN $\arg \min_{u}  \operatorname{Score} (u)$.
 \end{algorithmic}
 \end{algorithm}

\section{Numerical study}  \label{Sec:NumStudy}
In this section, we demonstrate our algorithms using both synthetic and real-world data.
The source codes are implemented by Python$3$.

\subsection{Synthetic data}
We first show the results of our methods with synthetic data.
We use three type of synthetic data;  1. generated with Gaussian distribution, 2. generated with Gamma distribution and 3. AR model.

\begin{figure}[htbp]
  \begin{center}
    \begin{tabular}{c}

      \begin{minipage}{0.33\hsize}
        \begin{center}
          \includegraphics[clip, width=6cm]{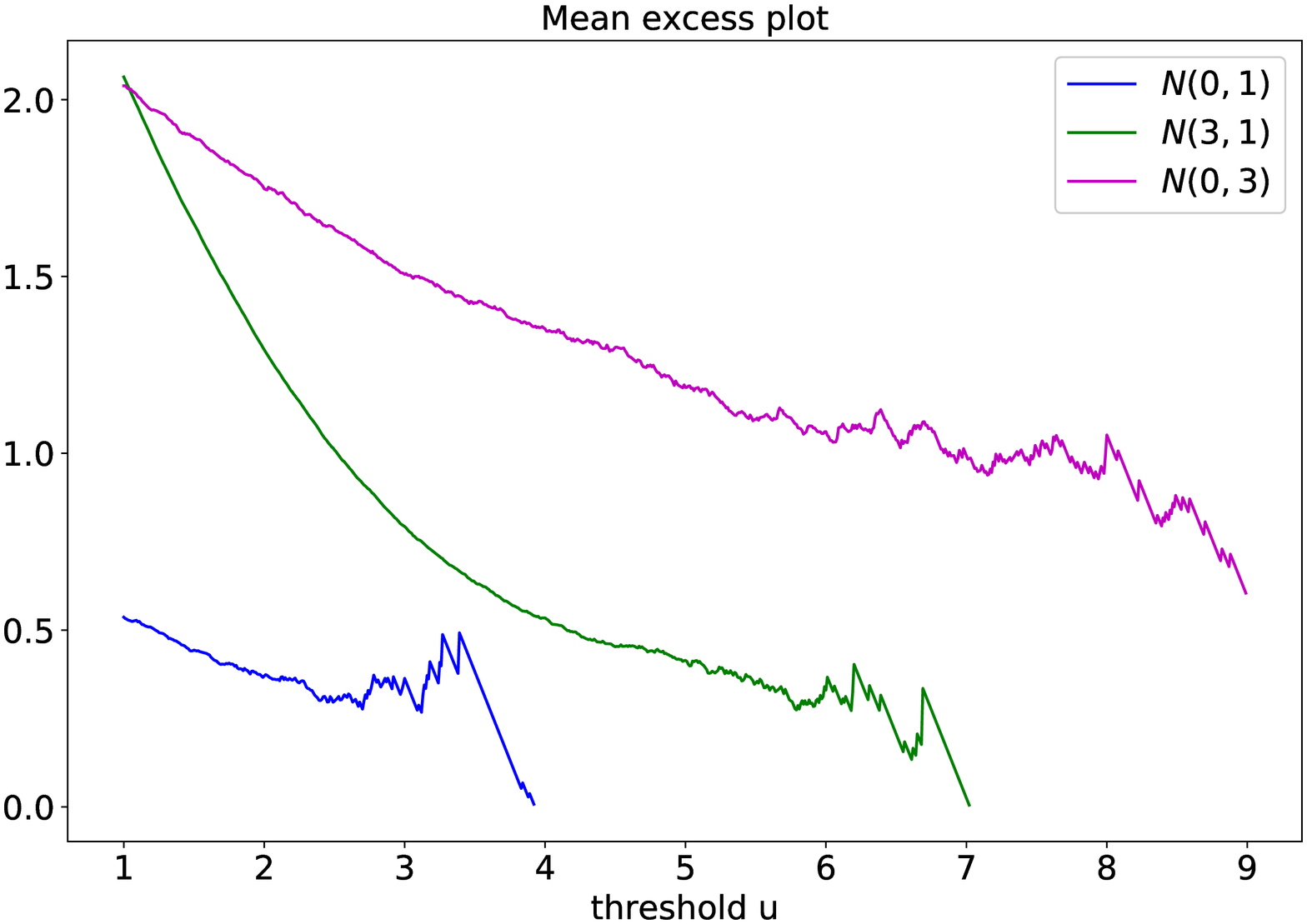}
        \end{center}
      \end{minipage}

      \begin{minipage}{0.33\hsize}
        \begin{center}
          \includegraphics[clip, width=6cm]{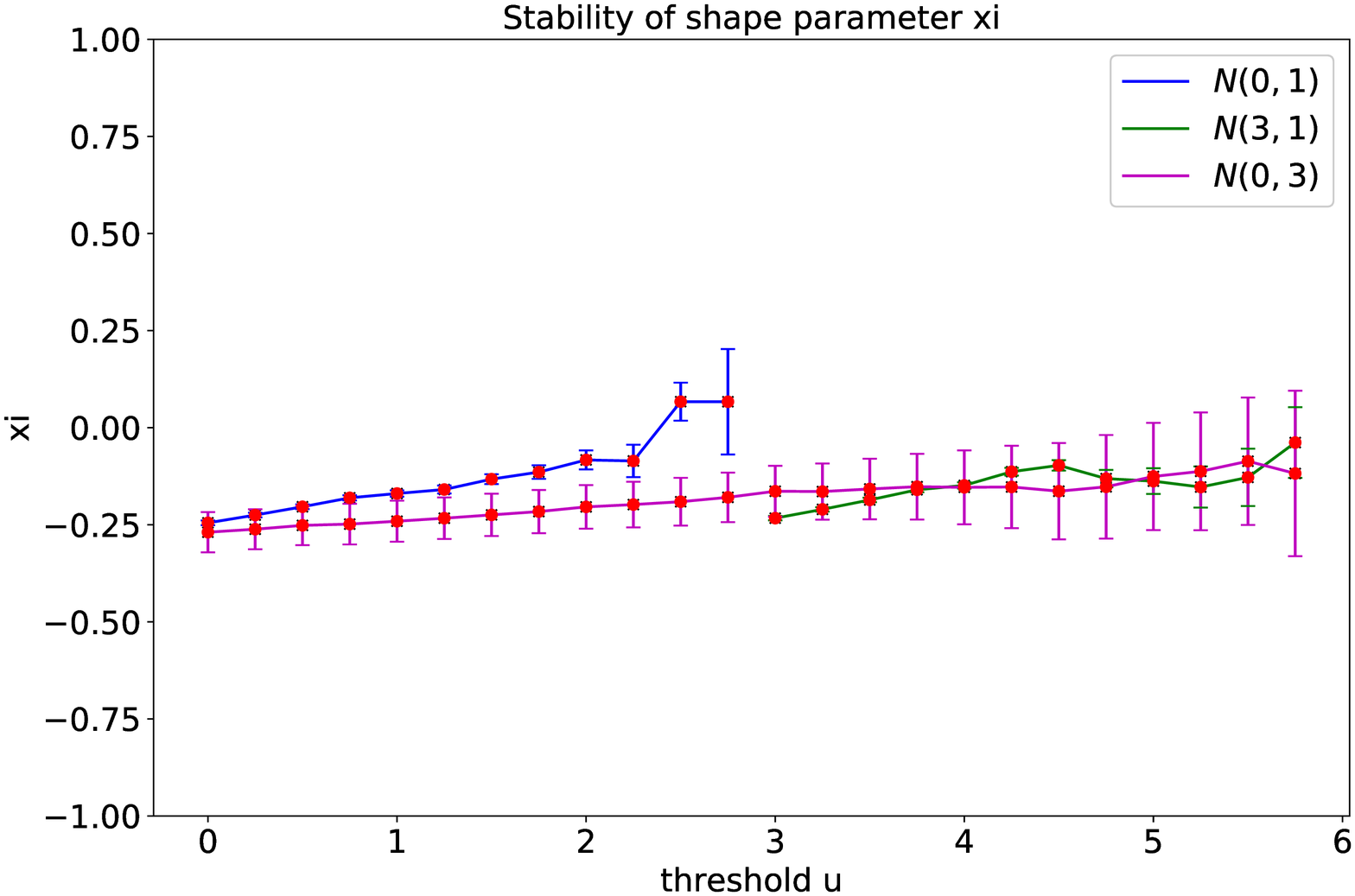}
        \end{center}
      \end{minipage}

      \begin{minipage}{0.33\hsize}
        \begin{center}
          \includegraphics[clip, width=6cm]{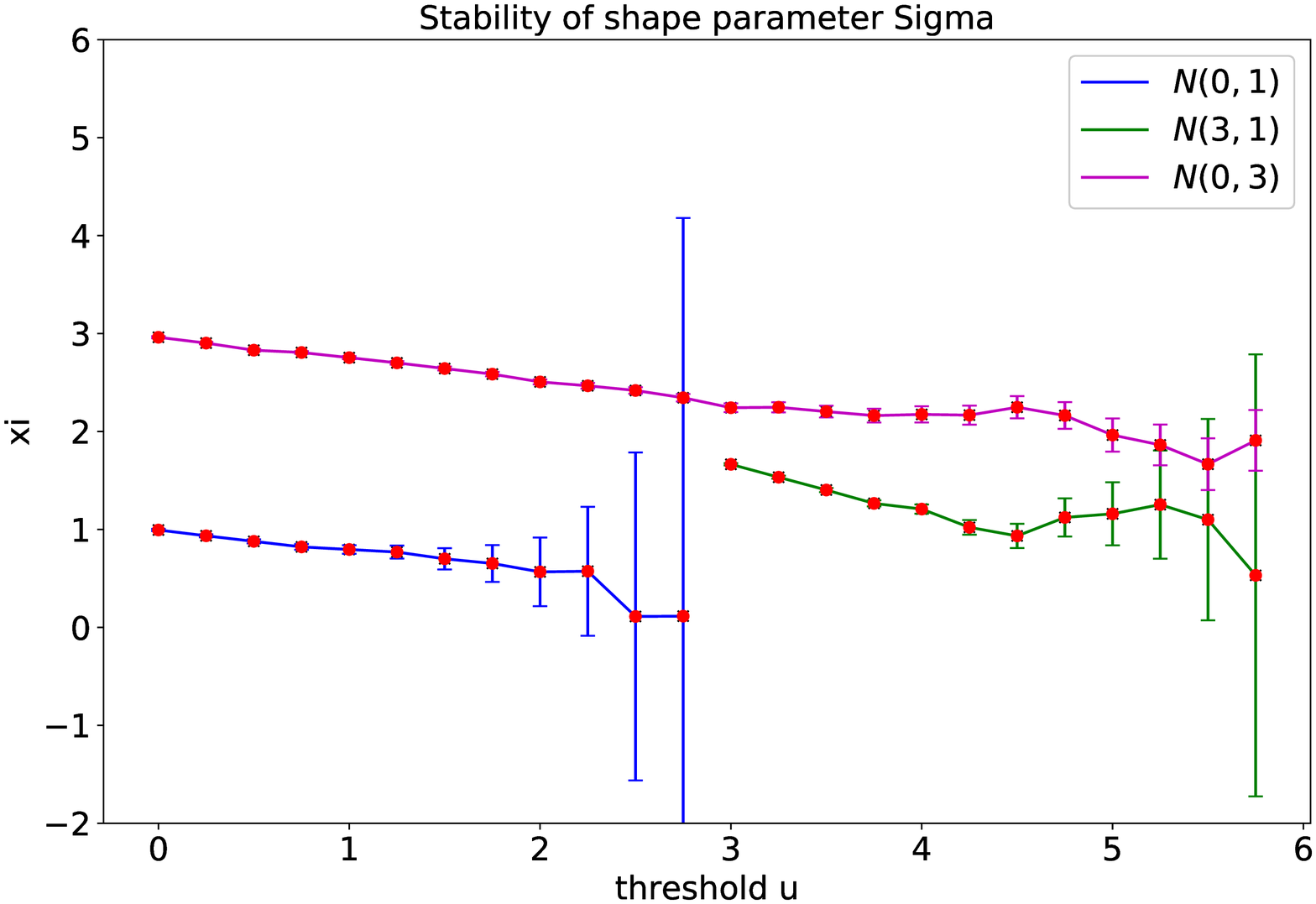}
        \end{center}
      \end{minipage}

    \end{tabular}
    \caption{Illustration of the mean excess plot, $\widehat{\xi}$ and $\widehat{\sigma}^{\ast}$ for a time series generated with $N(0,3)$.}
    \label{Fig:Syn1}
  \end{center}
\end{figure}

\begin{flushleft} {\bf Time series generated by Gaussian distribution} \end{flushleft}
In this experiment, we generate a time series with Gaussian distribution with time length $T=10000$.
We first select the search range of a threshold with the mean excess plots and the plots of $\widehat{\xi}, \widehat{\sigma}_{\ast}$.
Figure \ref{Fig:Syn1} shows illustrations of  the mean excess plots and the plots of $\widehat{\xi}, \widehat{\sigma}_{\ast}$.

From the viewpoints of linearity of the mean excess plot and  stability of $\widehat{\xi}$ and $\widehat{\sigma}^{\ast}$,
we select the search range of threshold $u$ as $(1.5, 4.0)$.
The following figures demonstrate a result of BO (in case of $N(0,3)$).
\begin{figure}[htbp]
  \begin{center}
    \begin{tabular}{c}

      \begin{minipage}{0.5\hsize}
        \begin{center}
          \includegraphics[clip, width=6cm]{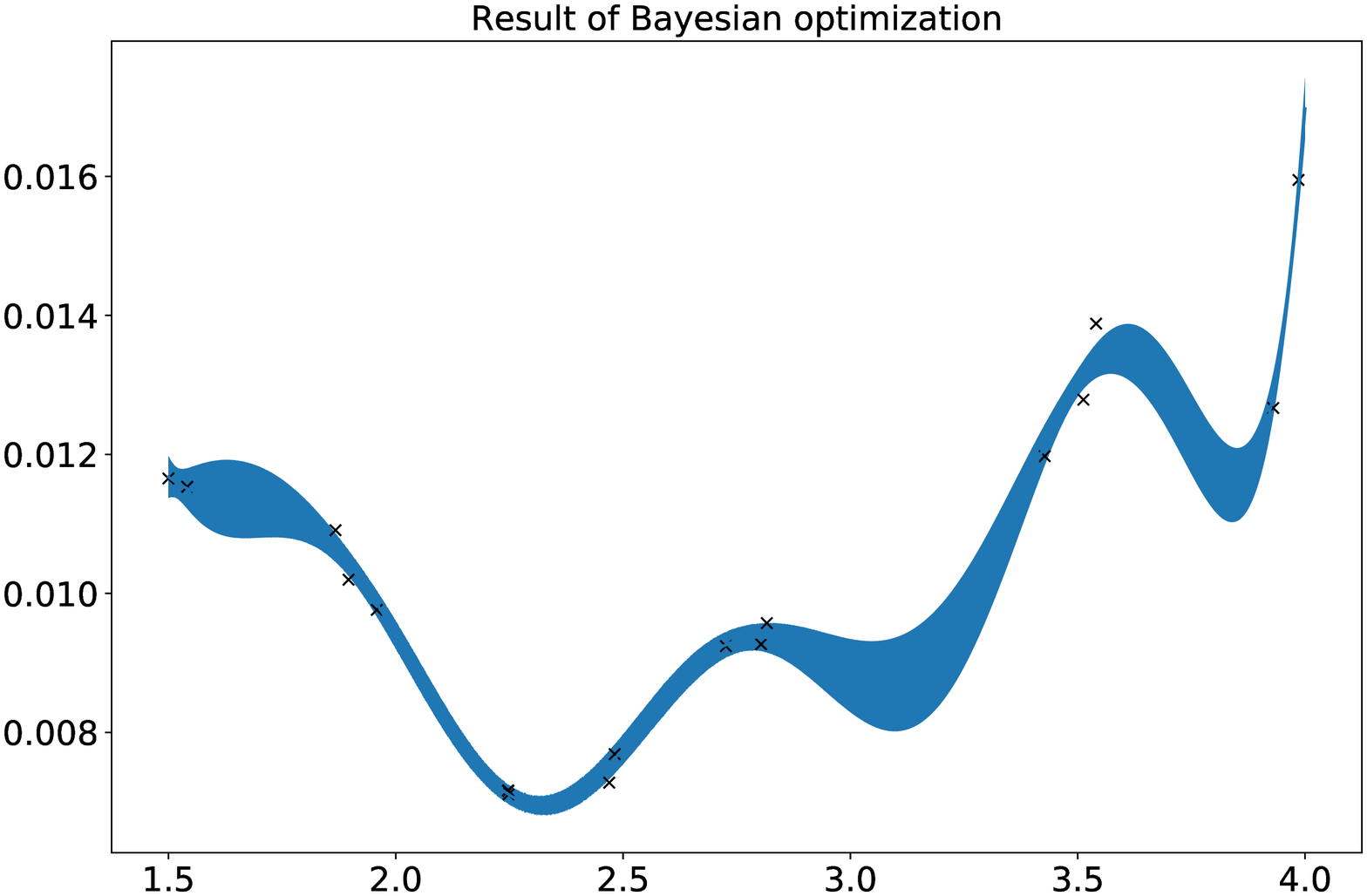}
        \end{center}
      \end{minipage}

      \begin{minipage}{0.5\hsize}
        \begin{center}
          \includegraphics[clip, width=6cm]{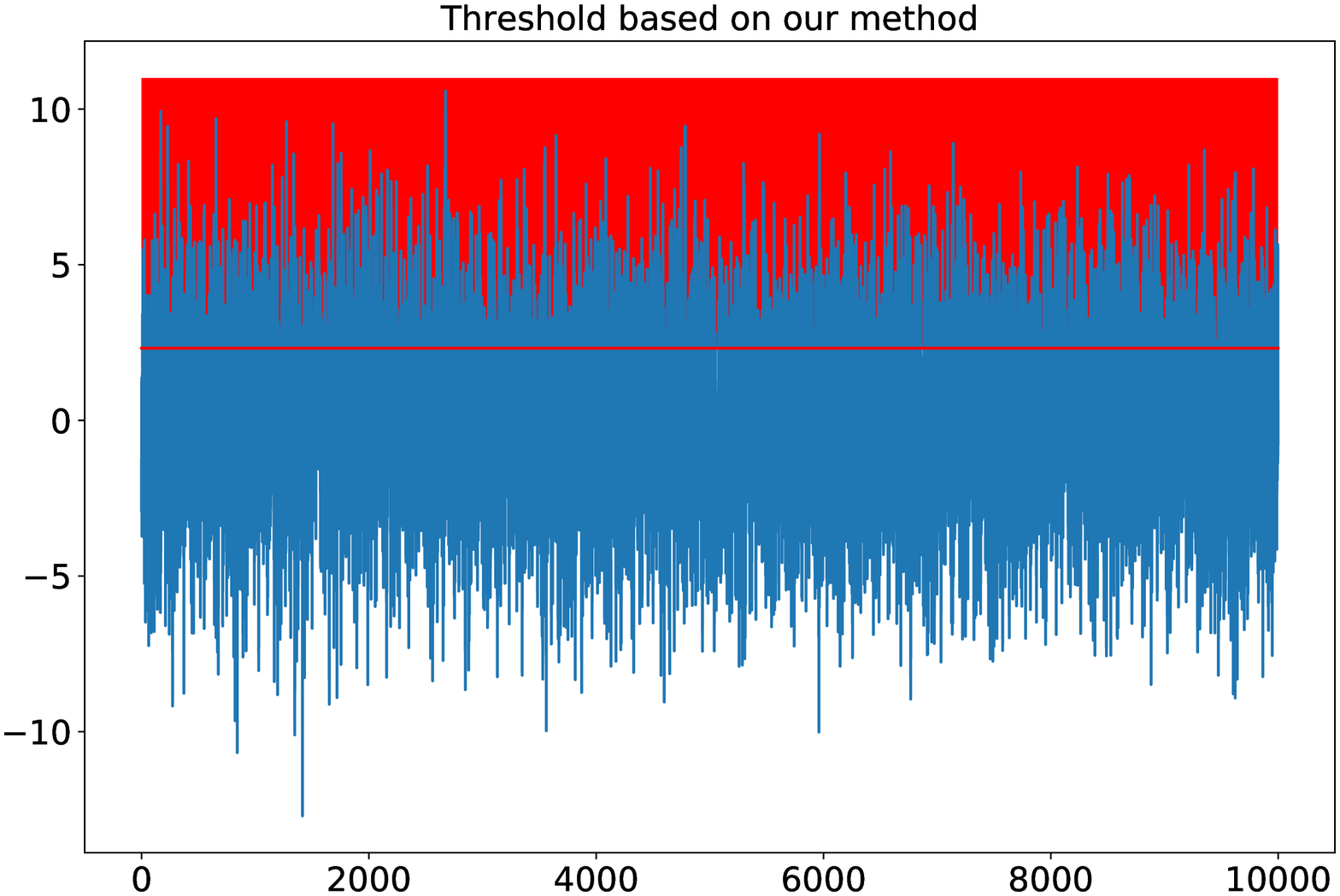}
        \end{center}
      \end{minipage}
    \end{tabular}
    \caption{Illustration of BO (left) and a choice of threshold (right). We choose the threshold $u$ as it attains the minimum of $\operatorname{score} (u)$.}
    %\label{Fig:Syn1}
  \end{center}
\end{figure}

We can check the fitness of the model as described in Section \ref{Sec:Check}.
Both probability and quantile plots are linear, all dots lie in 95 \% confidence bounds and density plot approximate the excess data.
So, the model is fitted to the GP model properly.

\medskip
The results of thresholds, $\hat{\xi}$ and $\hat{\sigma}$ are summarized in Table \ref{Table:BOGauss}.
\begin{table}[htbp]
  \caption{Results of BO for time series generated by Gaussian distribution}
  \label{Table:BOGauss}
\begin{center}
\begin{tabular}{|c|c|c|c|c|c|} %\label{Table:BOGauss}
\hline Probability &  domain (of BO) & Trial & threshold   & $\hat{\xi}$ & $\widehat{\sigma}$  \\
\hline$N(0,1)$     & (1.0,2.5) & 1 & 1.198 &  -0.0921 & 0.5332 \\
\cline { 3 - 6}    &           &2  & 1.027 &  -0.1502 & 0.5768 \\
\cline { 3 - 6 }   &           & 3 & 1.026 & -0.1749 & 0.6037 \\
\hline$N(0,3)$    & (1.5,4.0)  & 1 & 2.800 & -0.1504 & 1.8072 \\
\cline { 3 - 6 } &             &  2 & 2.894  & -0.1624 & 1.8980 \\
\cline { 3 - 6 } &             & 3 & 2.321 & -0.2167 & 2.0609 \\
\hline$N(3,1)$   & (4.0,5.5)   & 1 & 4.008 & -0.1317 & 0.5711 \\
\cline { 3 - 6} &              & 2 & 4.360 & -0.1052 & 0.4983 \\
\cline { 3 - 6 } &             & 3 & 4.015 & -0.1902 & 0.6033 \\
\hline
\end{tabular}
\end{center}
\end{table}

\begin{flushleft} {\bf Time series generated by gamma distribution} \end{flushleft}
Similarly as the case of Gaussian distribution, we demonstrate our approach for time series generated by Gamma distribution.

\begin{figure}[htbp]
  \begin{center}
    \begin{tabular}{c}

      \begin{minipage}{0.33\hsize}
        \begin{center}
          \includegraphics[clip, width=6cm]{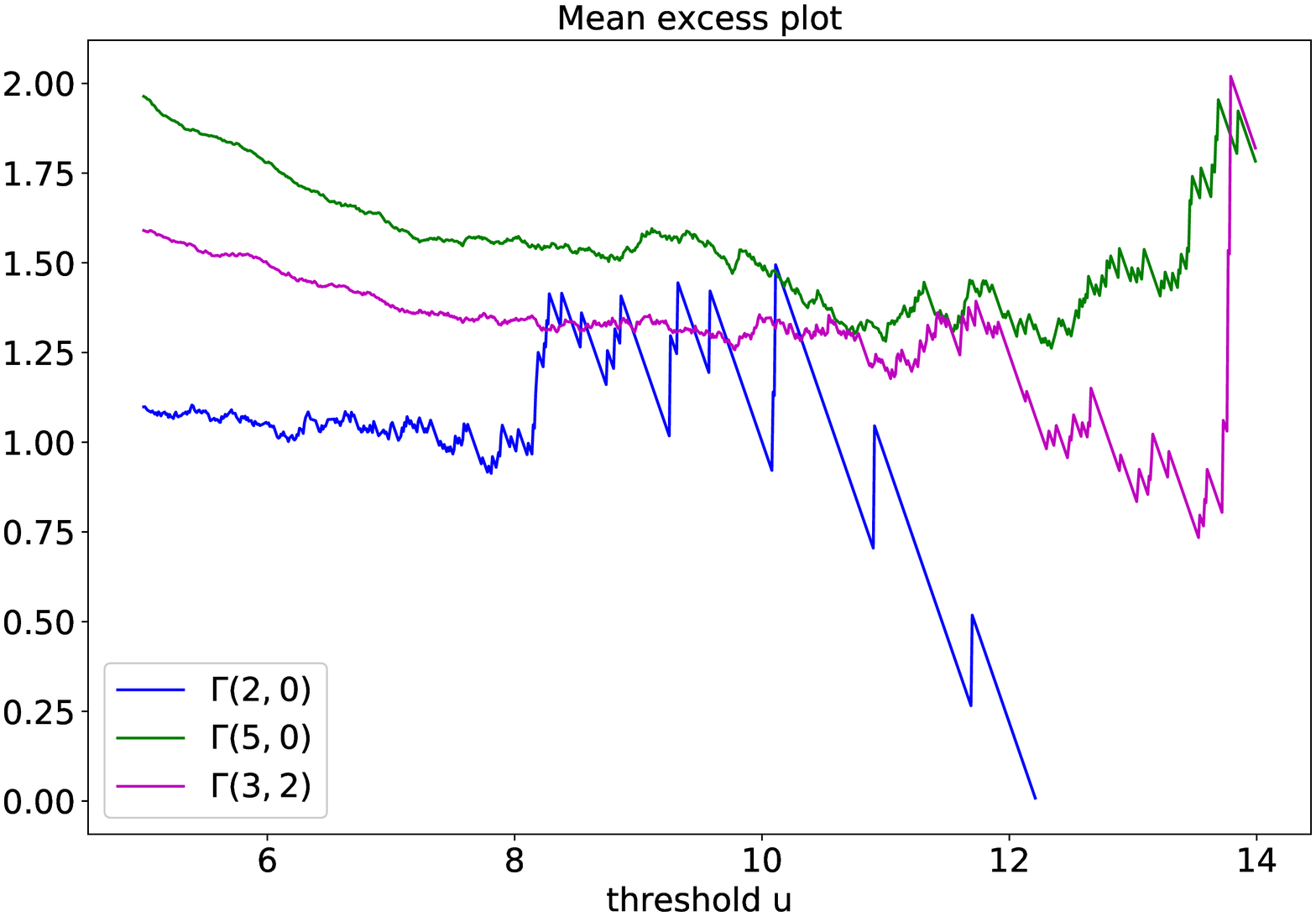}
        \end{center}
      \end{minipage}

      \begin{minipage}{0.33\hsize}
        \begin{center}
          \includegraphics[clip, width=6cm]{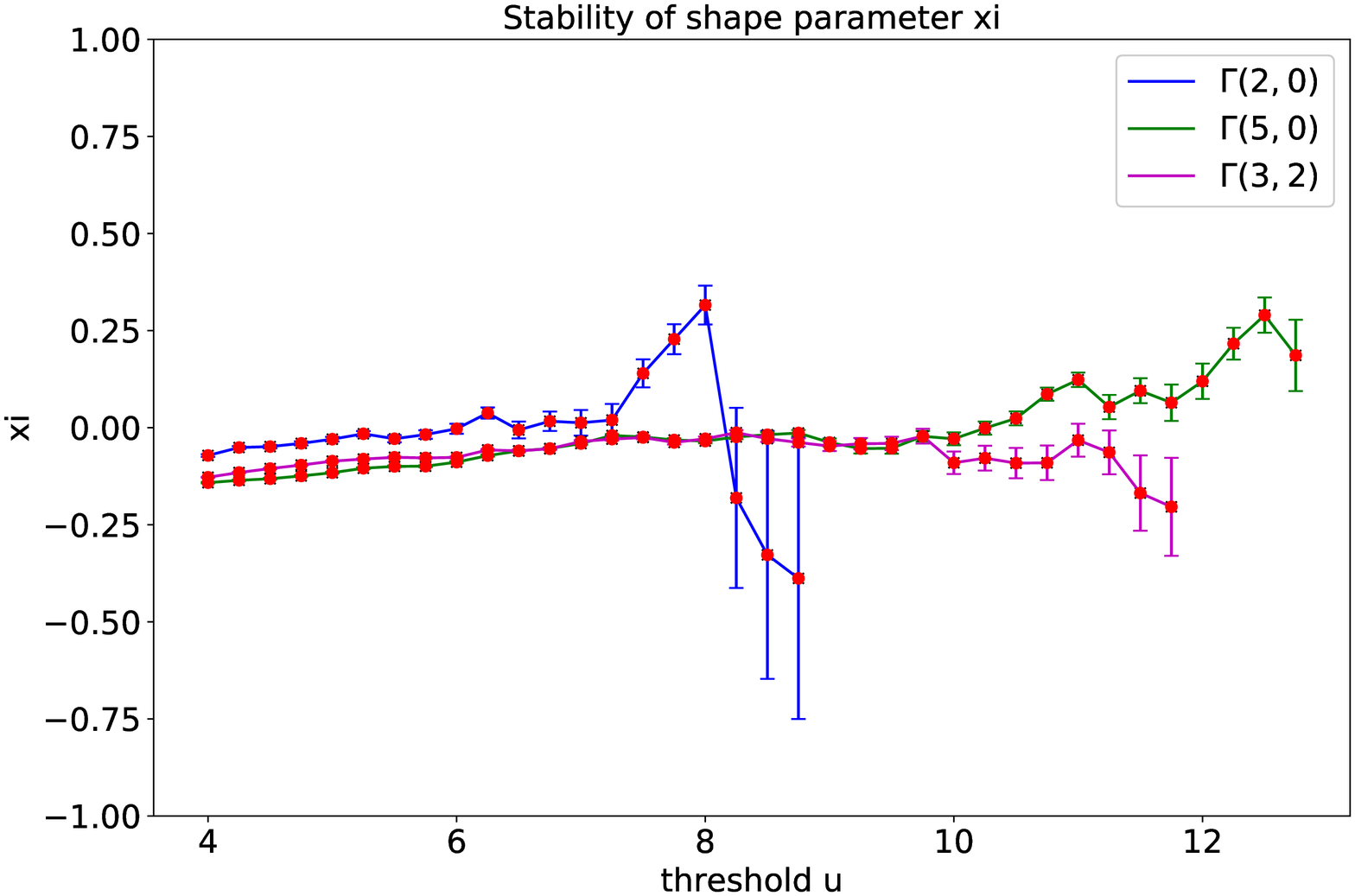}
        \end{center}
      \end{minipage}

      \begin{minipage}{0.33\hsize}
        \begin{center}
          \includegraphics[clip, width=6cm]{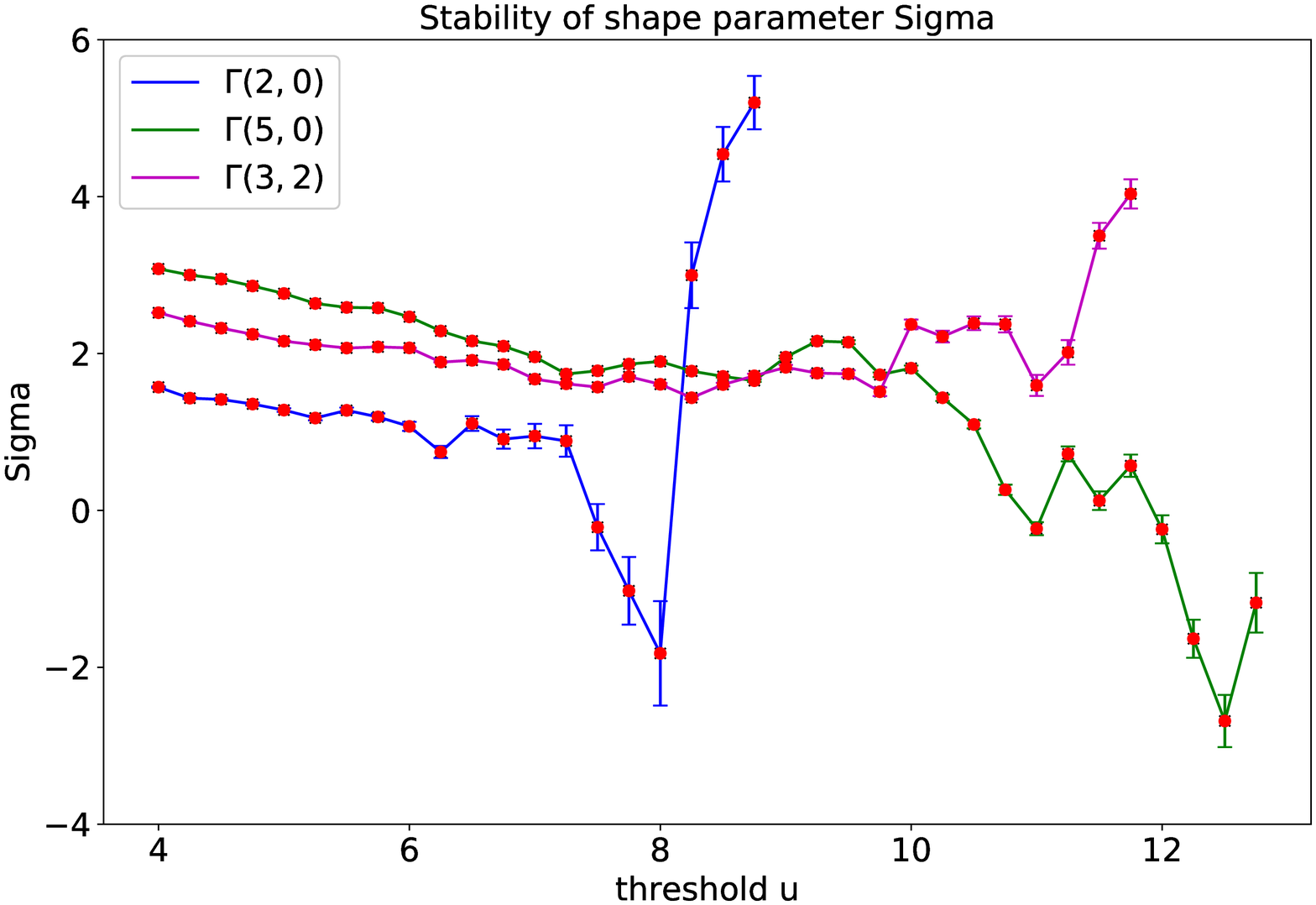}
        \end{center}
      \end{minipage}

    \end{tabular}
    \caption{Illustration of the mean excess plot, $\widehat{\xi}$ and $\widehat{\sigma}^{\ast}$ for a time series generated with gamma distribution.}
    \label{Fig:SynGam}
  \end{center}
\end{figure}

We first select the search range of a threshold with the mean excess plots and the plots of $\widehat{\xi}, \widehat{\sigma}_{\ast}$
 (see Fig \ref{Fig:SynGam}).

\begin{figure}[htbp]
  \begin{center}
    \begin{tabular}{c}

      \begin{minipage}{0.5\hsize}
        \begin{center}
          \includegraphics[clip, width=6cm]{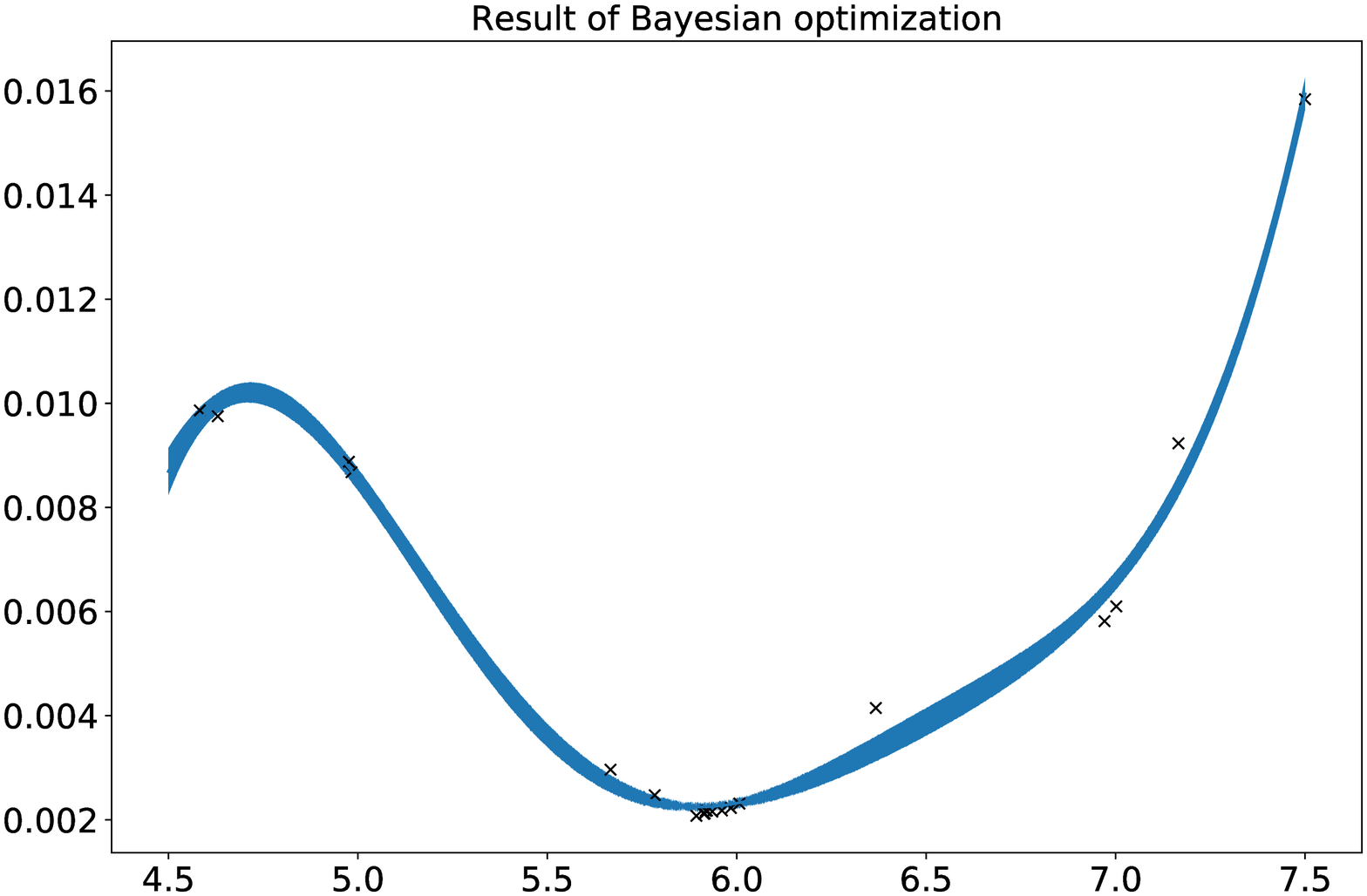}
        \end{center}
      \end{minipage}

      \begin{minipage}{0.5\hsize}
        \begin{center}
          \includegraphics[clip, width=6cm]{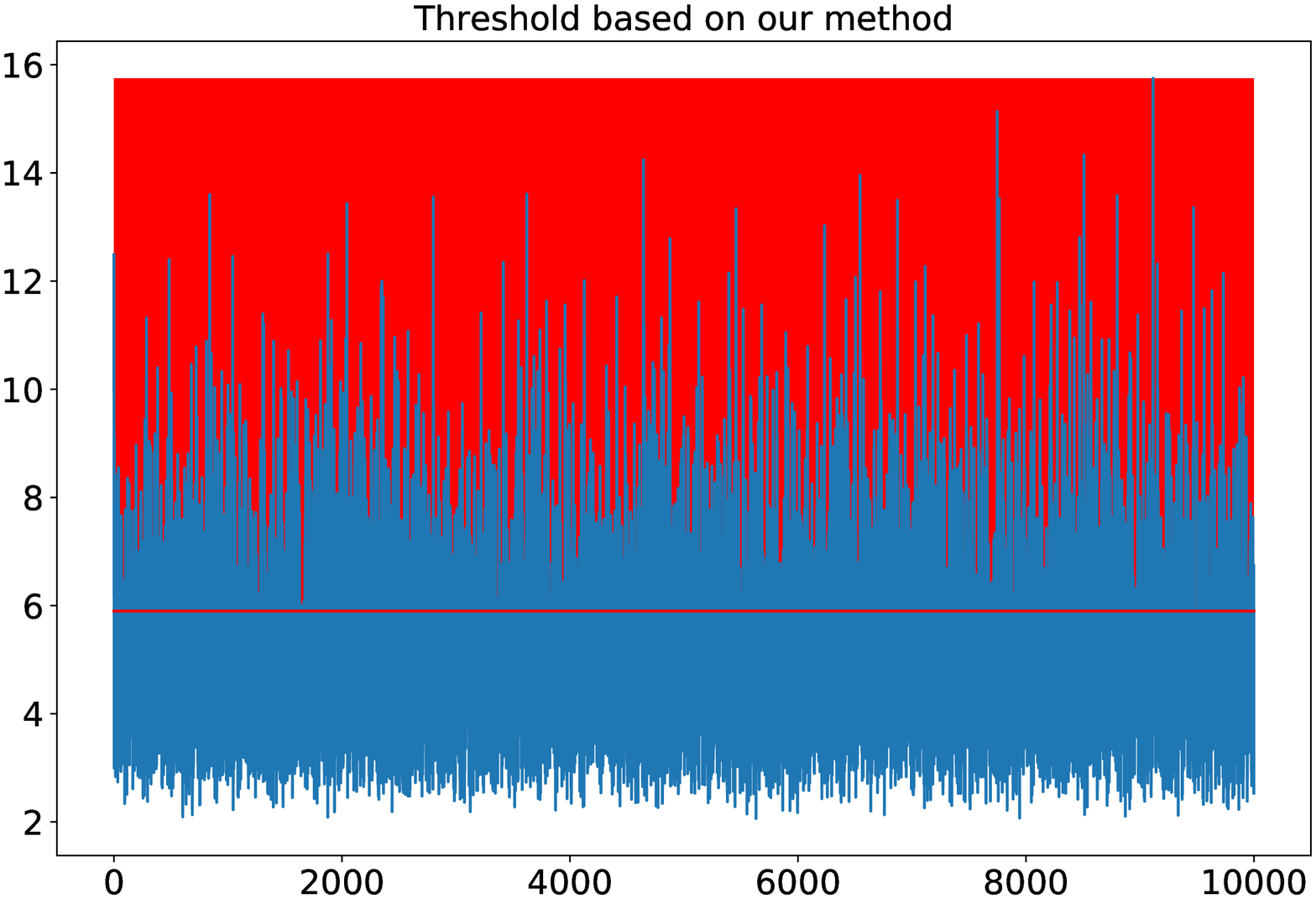}
        \end{center}
      \end{minipage}
    \end{tabular}
    \caption{Illustration of BO (left) and a choice of threshold (right). We choose the threshold $u$ as it attains the minimum of $\operatorname{score} (u)$.}
    \label{Fig:BO_Gam1}
  \end{center}
\end{figure}

\begin{figure}[htbp]
  \begin{center}
    \begin{tabular}{c}
      \begin{minipage}{0.5\hsize}
        \begin{center}
          \includegraphics[clip, width=6cm]{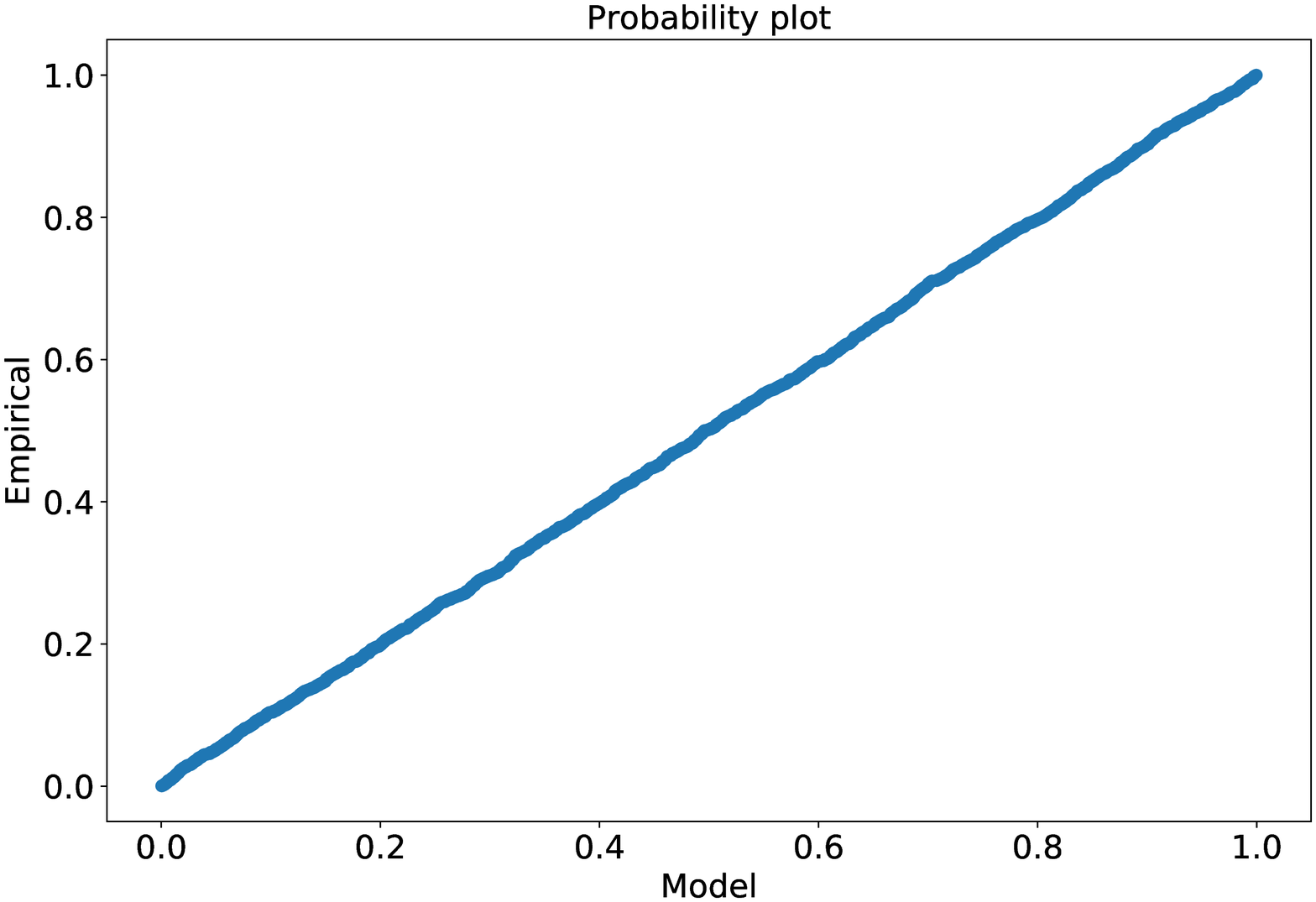}
        \end{center}
      \end{minipage}

      \begin{minipage}{0.5\hsize}
        \begin{center}
          \includegraphics[clip, width=6cm]{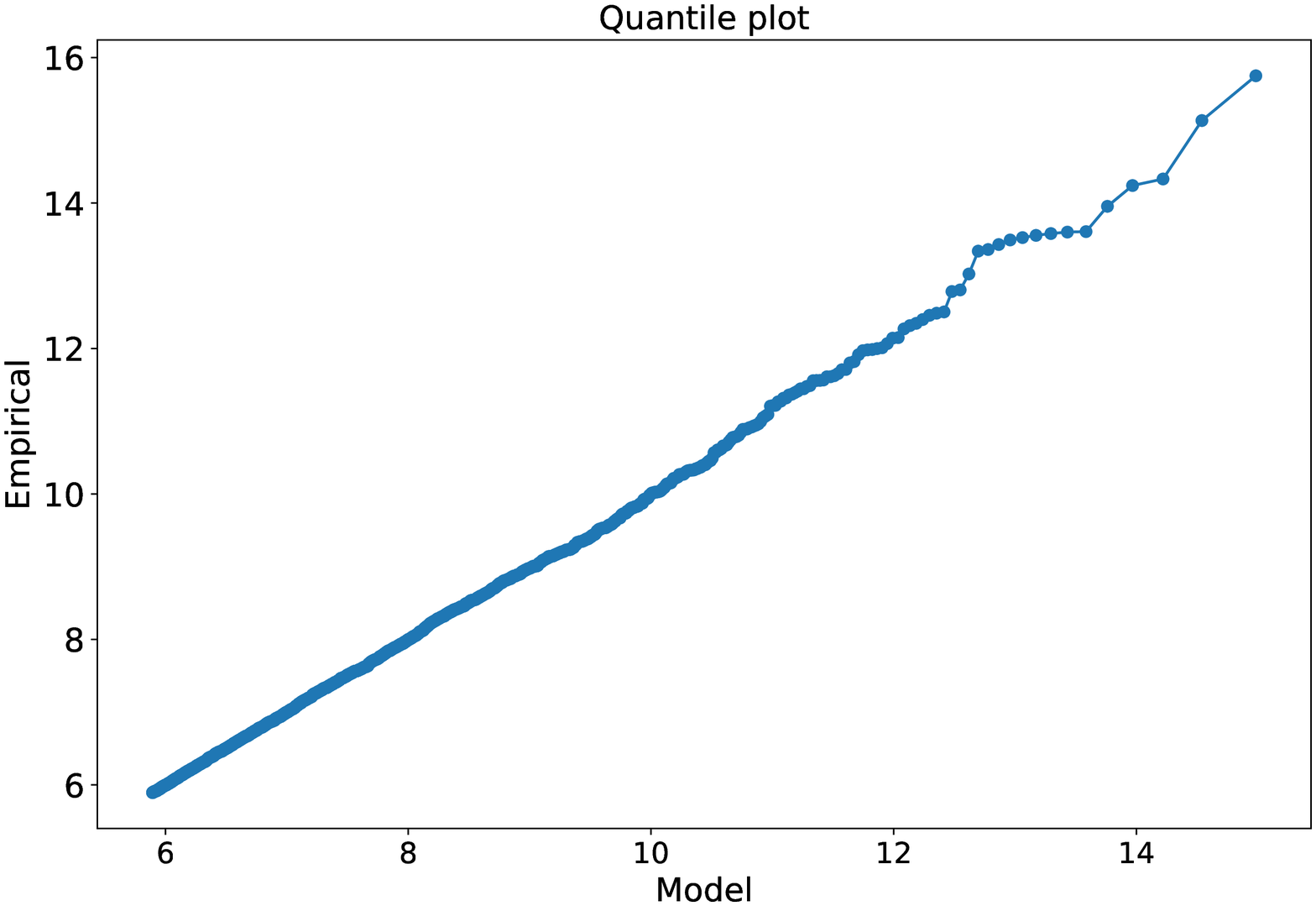}
        \end{center}
      \end{minipage}\\
      \begin{minipage}{0.5\hsize}
        \begin{center}
          \includegraphics[clip, width=6cm]{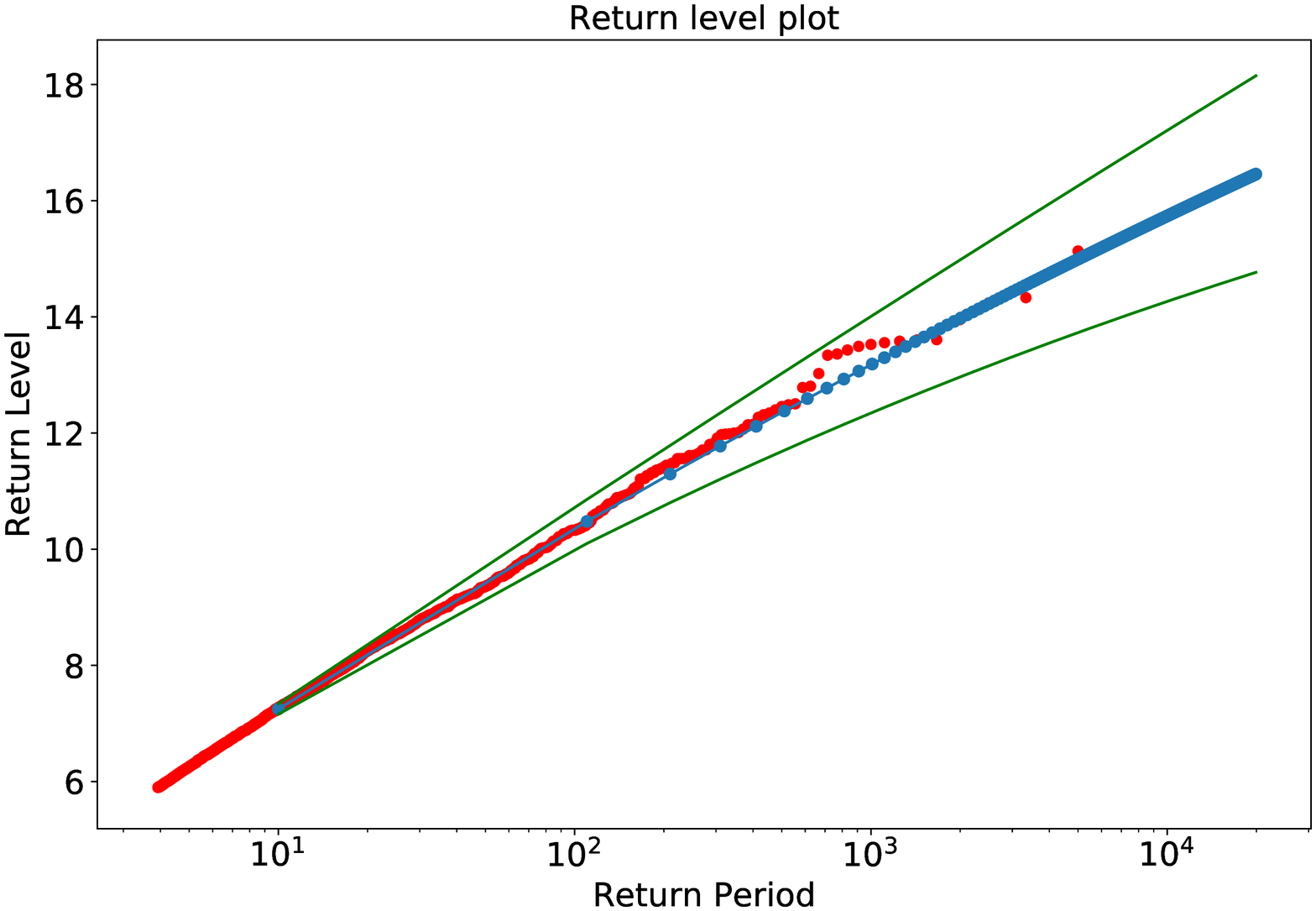}
        \end{center}
      \end{minipage}

      \begin{minipage}{0.5\hsize}
        \begin{center}
          \includegraphics[clip, width=6cm]{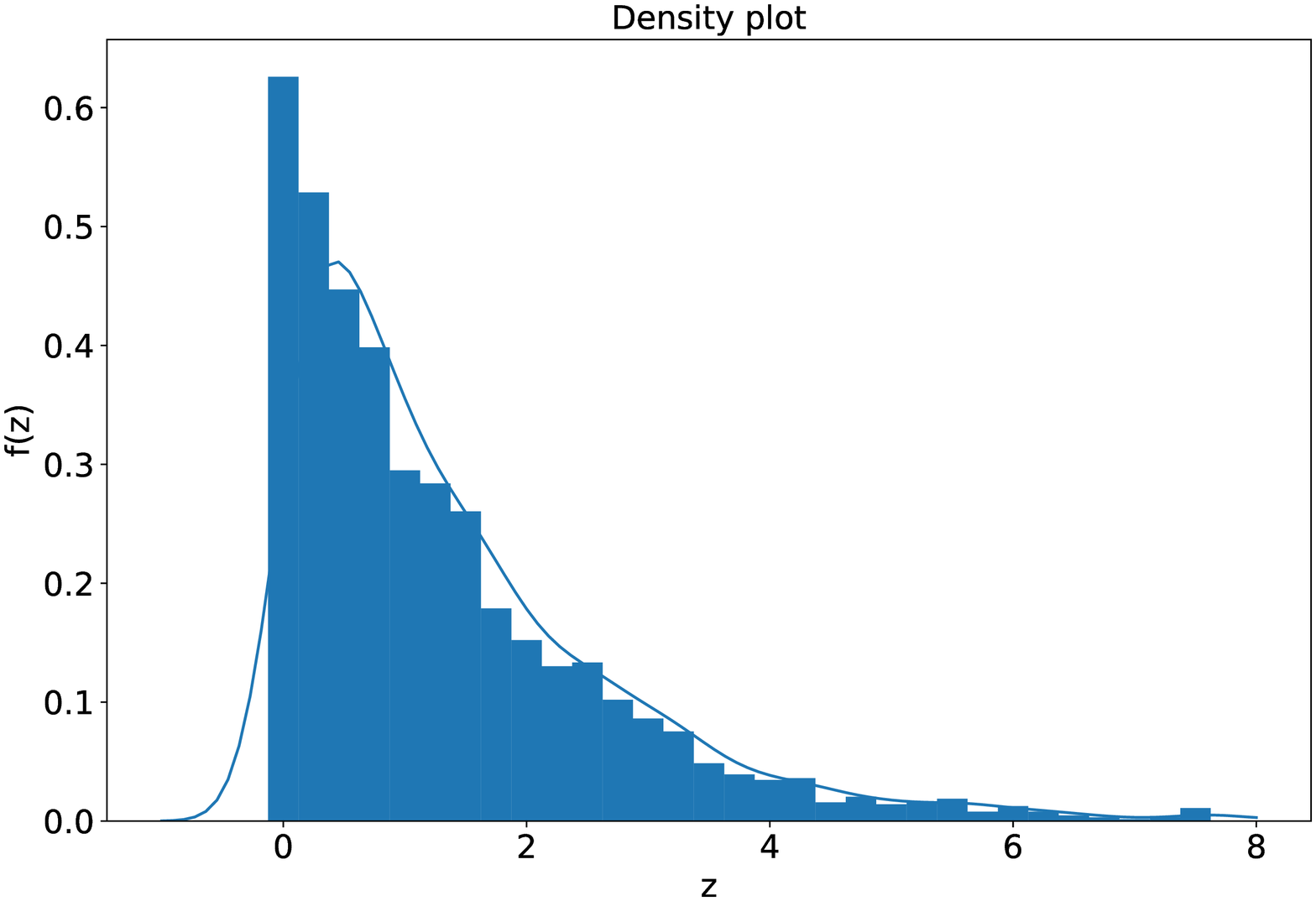}
        \end{center}
      \end{minipage}\\

    \end{tabular}
    \caption{Illustrations of probability plot, quantile plot, return level plot and density plot for the $u$ selected above.}
    \label{Fig:Check_Gam1}
  \end{center}
\end{figure}

We select the search range of threshold $u$ based on linearity of the mean excess plot and stability of $\widehat{\xi}$ and $\widehat{\sigma}^{\ast}$.
Then, we apply BO and search the value of threshold.
Figures  \ref{Fig:BO_Gam1} and  \ref{Fig:Check_Gam1} illustrate the result of BO and fitness to GP model described in Section \ref{Sec:Check}.
We can confirm both probability and quantile plots are linear, return level plots lie in 95 \% confidence bounds and density plot approximate the excess data,
and hence the model is fitted to the GP model properly.

The results of threshold,$\hat{\xi}$ and $\hat{\sigma}$ are summarized in Table \ref{Table:Gamma}.

\begin{table}[htbp]
    \centering
    \caption{Results of BO for time series generated by gamma distribution}
    \begin{tabular}{|c|c|c|c|c|c|}
    \hline Probability & domain (of BO) & Trial & threshold & $\hat{\xi}$ & $\widehat{\sigma}$ \\
    \hline {$\Gamma(2,0)$} & (2.5,5.0)   & 1    & 3.646      & -0.1107    & 1.3753 \\
    \cline { 3 - 6 } &     & 2 & 3.366 & -0.0322 & 1.2305 \\
    \cline { 3 - 6 } &     & 3 & 2.956 & -0.042 & 1.3255 \\
    \hline {$\Gamma (5,0)$} &  {(6.0,10.0)}    & 1 & 7.379 & -0.0257 & 1.6381 \\
    \cline { 3 - 6 }     &       &  2     & 7.404 & -0.085 & 1.7583 \\
    \cline { 3 - 6 } & & 3 & 7.231 & -0.0988 & 1.7752 \\
    \hline {$\Gamma(3,2)$} &  {(4.5,7.5)} & 1 & 5.898 & -0.0418 & 1.4715 \\
    \cline { 3 - 6 } &   &2 & 6.655 & -0.0505 & 1.4839 \\
    \cline { 3 - 6 } &   & 3 & 7.382 & -0.0556 & 1.4277 \\
    \hline
\end{tabular}
 \label{Table:Gamma}
\end{table}

\subsection{Real-world data}
In this section, we demonstrate our methods for rainfall data of Japanese cities.
The data is available at Japan Meteorological Agency Website\footnote{https://www.jma.go.jp/jma/index.html}.
For this demonstration, we use rainfall data of cities in Japan, Tokyo, Osaka and Fukuoka, between January 1st 2000 and December 31th 2019.

\medskip
Similarly in the previous section, we first select the   search range of a threshold.
Fig \label{Fig:Rain} illustrates the mean excess plots and the plots of $\widehat{\xi}, \widehat{\sigma}_{\ast}$.

\begin{figure}[htbp]
  \begin{center}
    \begin{tabular}{c}

      \begin{minipage}{0.33\hsize}
        \begin{center}
          \includegraphics[clip, width=6cm]{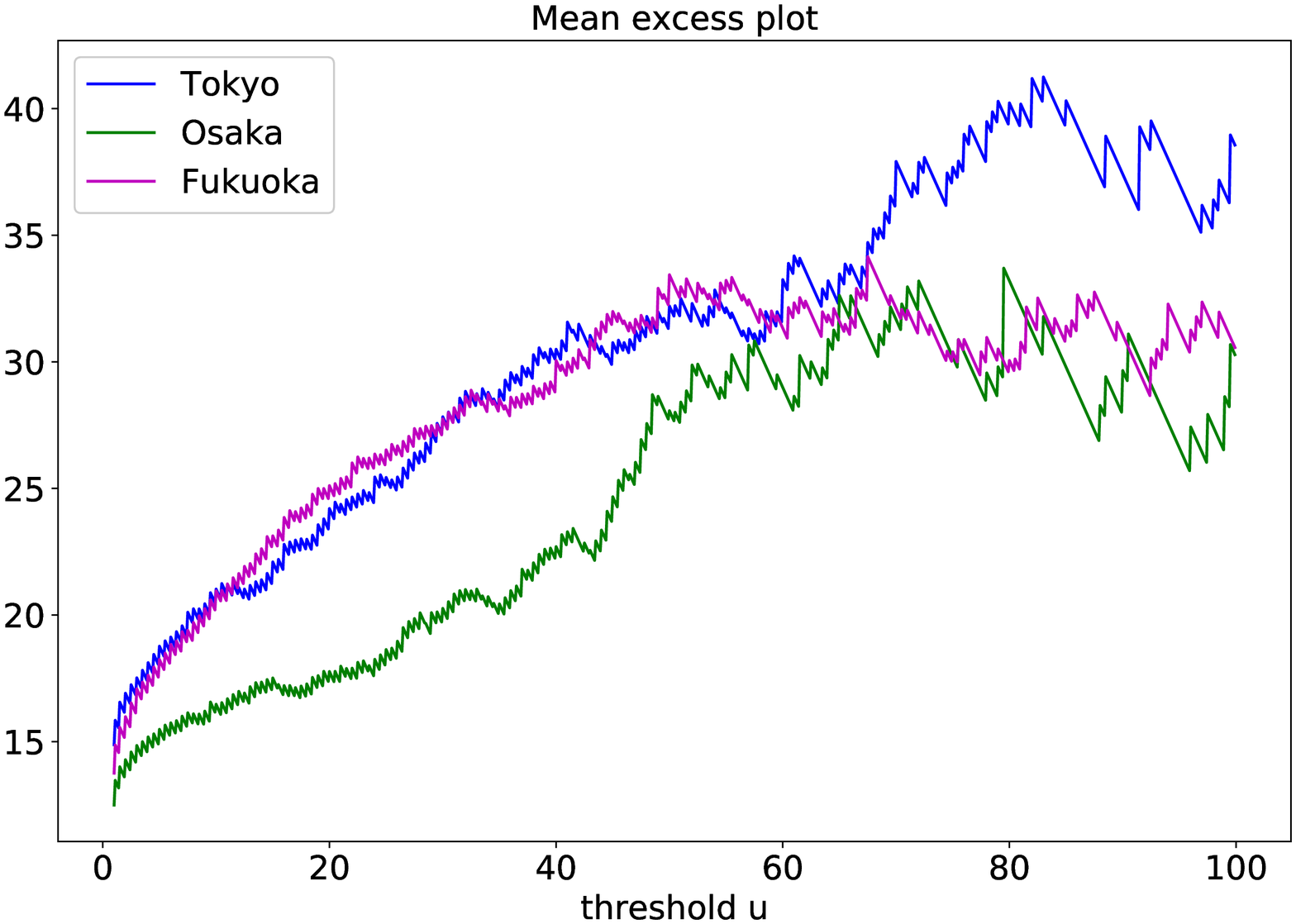}
        \end{center}
      \end{minipage}

      \begin{minipage}{0.33\hsize}
        \begin{center}
          \includegraphics[clip, width=6cm]{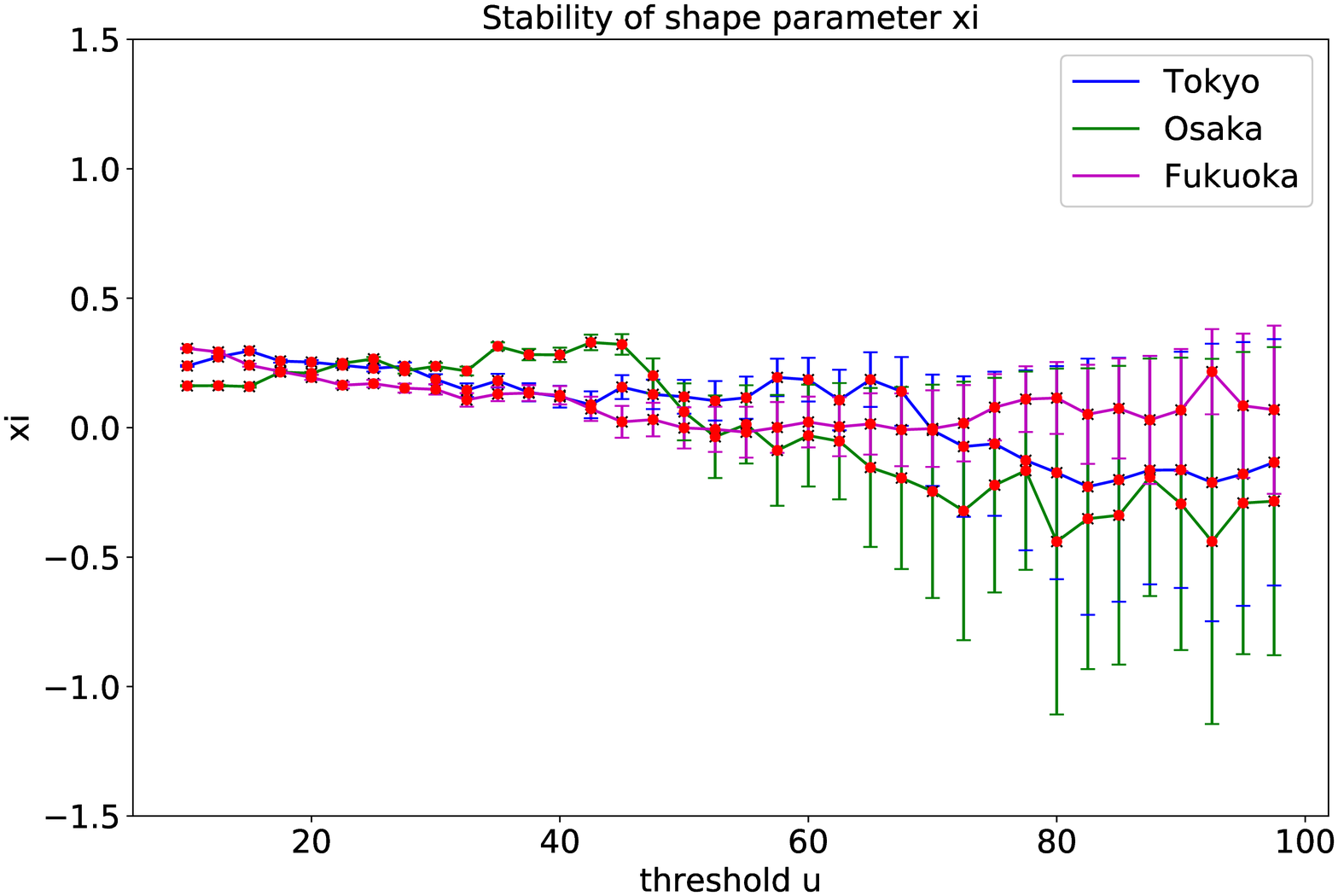}
        \end{center}
      \end{minipage}

      \begin{minipage}{0.33\hsize}
        \begin{center}
          \includegraphics[clip, width=6cm]{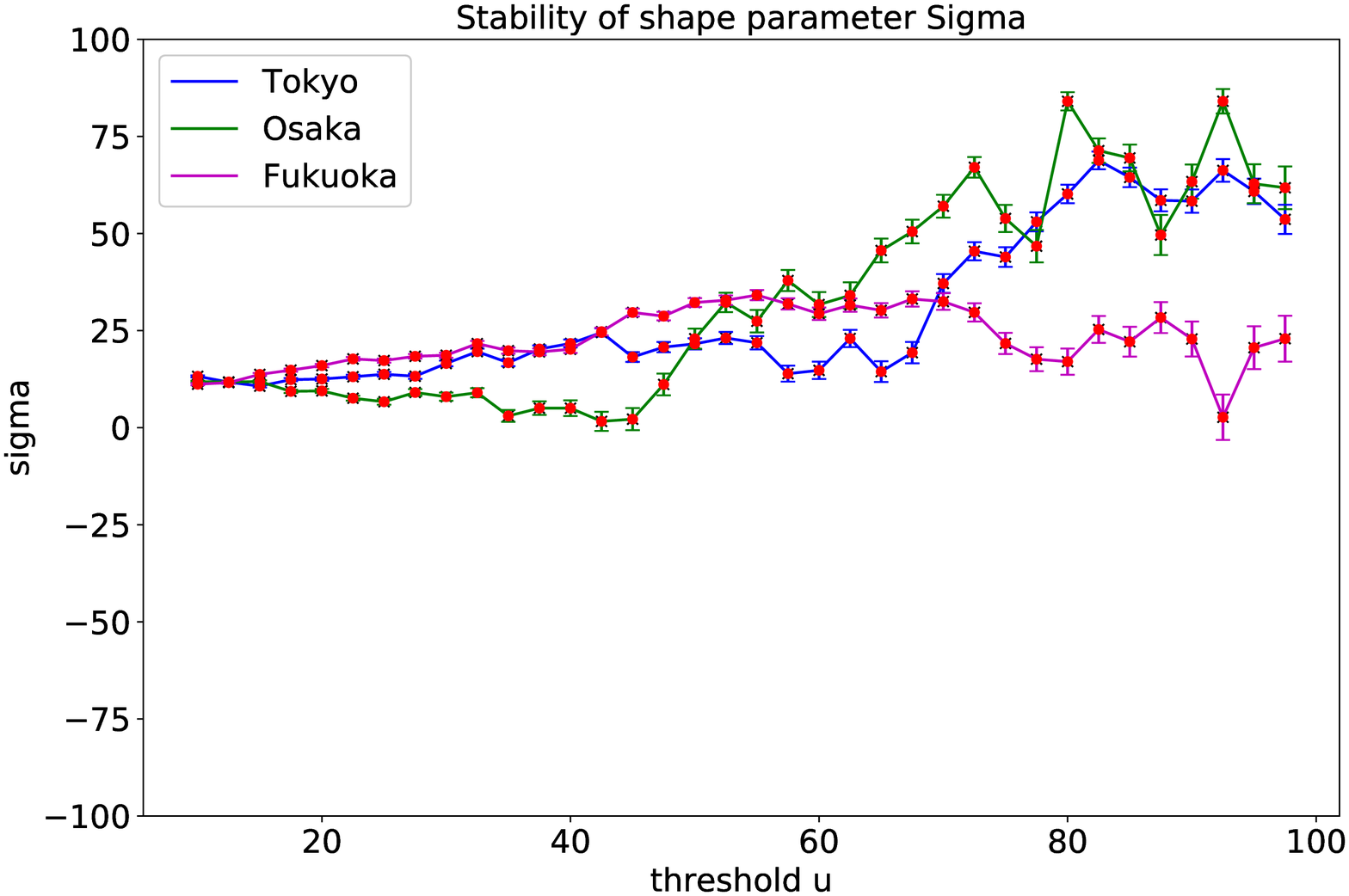}
        \end{center}
      \end{minipage}

    \end{tabular}
    \caption{Illustration of the mean excess plot, $\widehat{\xi}$ and $\widehat{\sigma}^{\ast}$ of rainfall data at Tokyo, Osaka and Fukuoka.}
    \label{Fig:Rain}
  \end{center}
\end{figure}

\begin{figure}[htbp]
  \begin{center}
    \begin{tabular}{c}

      \begin{minipage}{0.5\hsize}
        \begin{center}
          \includegraphics[clip, width=6cm]{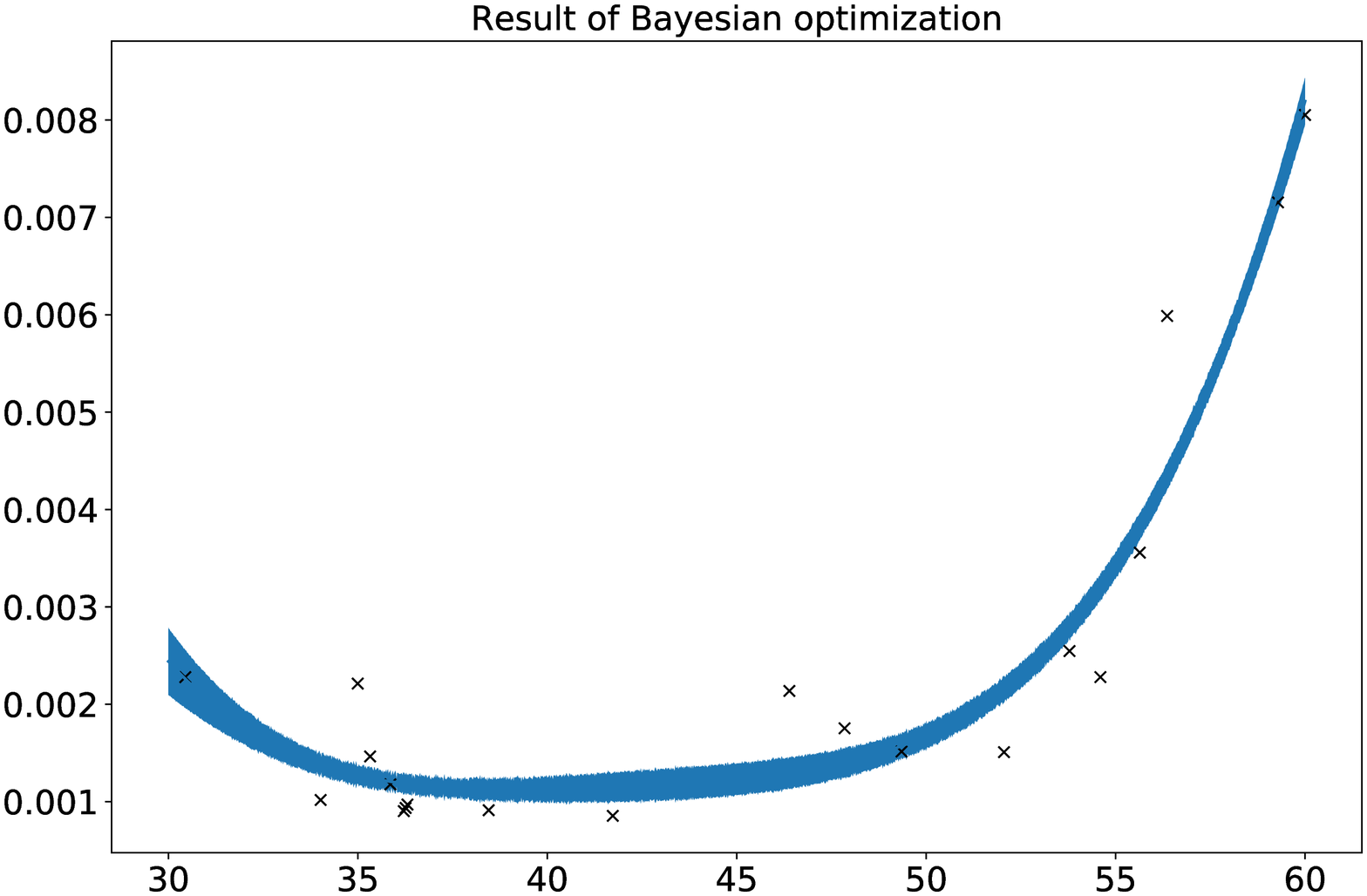}
        \end{center}
      \end{minipage}

      \begin{minipage}{0.5\hsize}
        \begin{center}
          \includegraphics[clip, width=6cm]{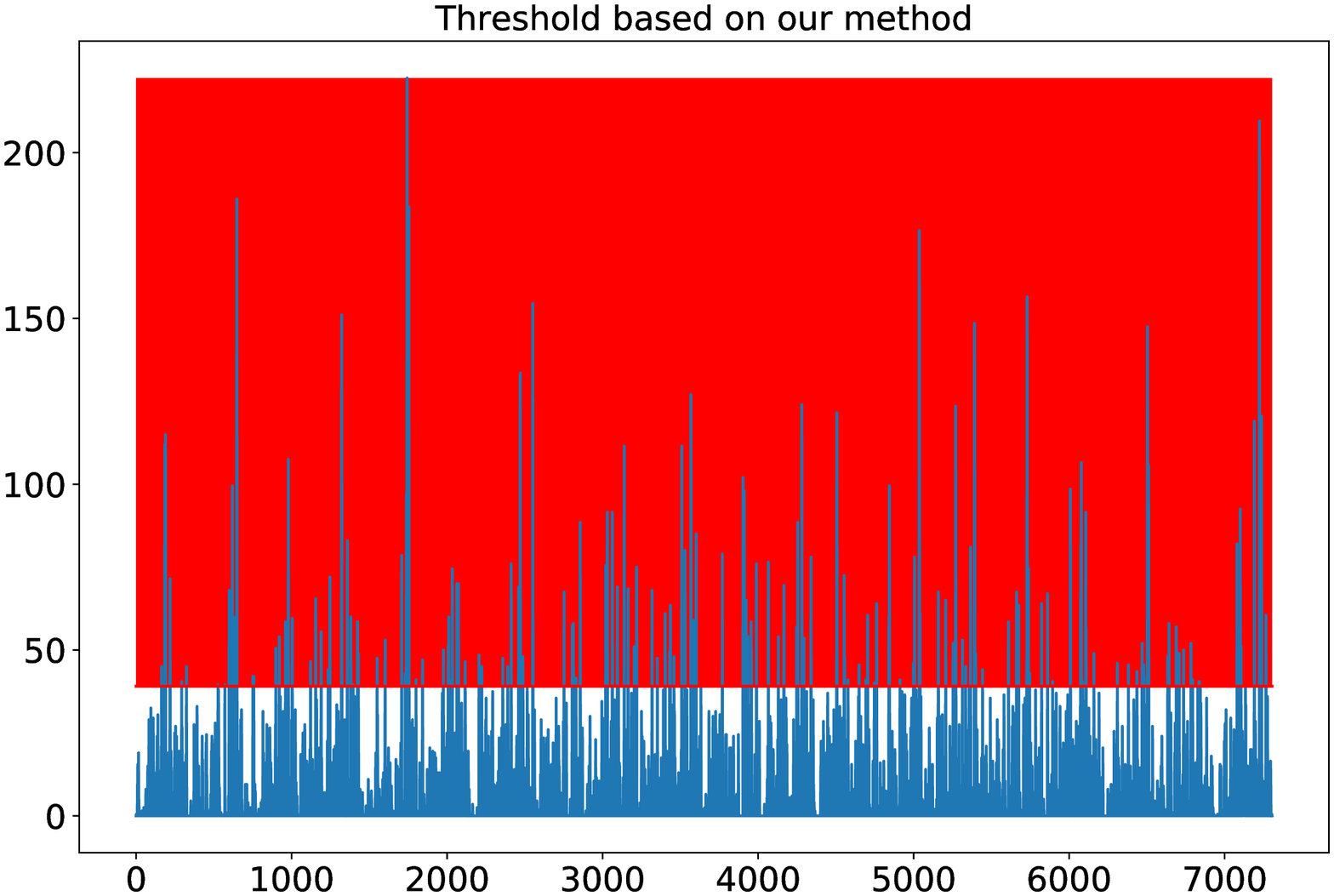}
        \end{center}
      \end{minipage}
    \end{tabular}
    \caption{Illustration of BO (left) and a choice of threshold (right). We choose the threshold $u$ as it attains the minimum of $\operatorname{score} (u)$.}
    \label{Fig:BO_Real}
  \end{center}
\end{figure}

\begin{figure}[htbp]
  \begin{center}
    \begin{tabular}{c}
      \begin{minipage}{0.5\hsize}
        \begin{center}
          \includegraphics[clip, width=6cm]{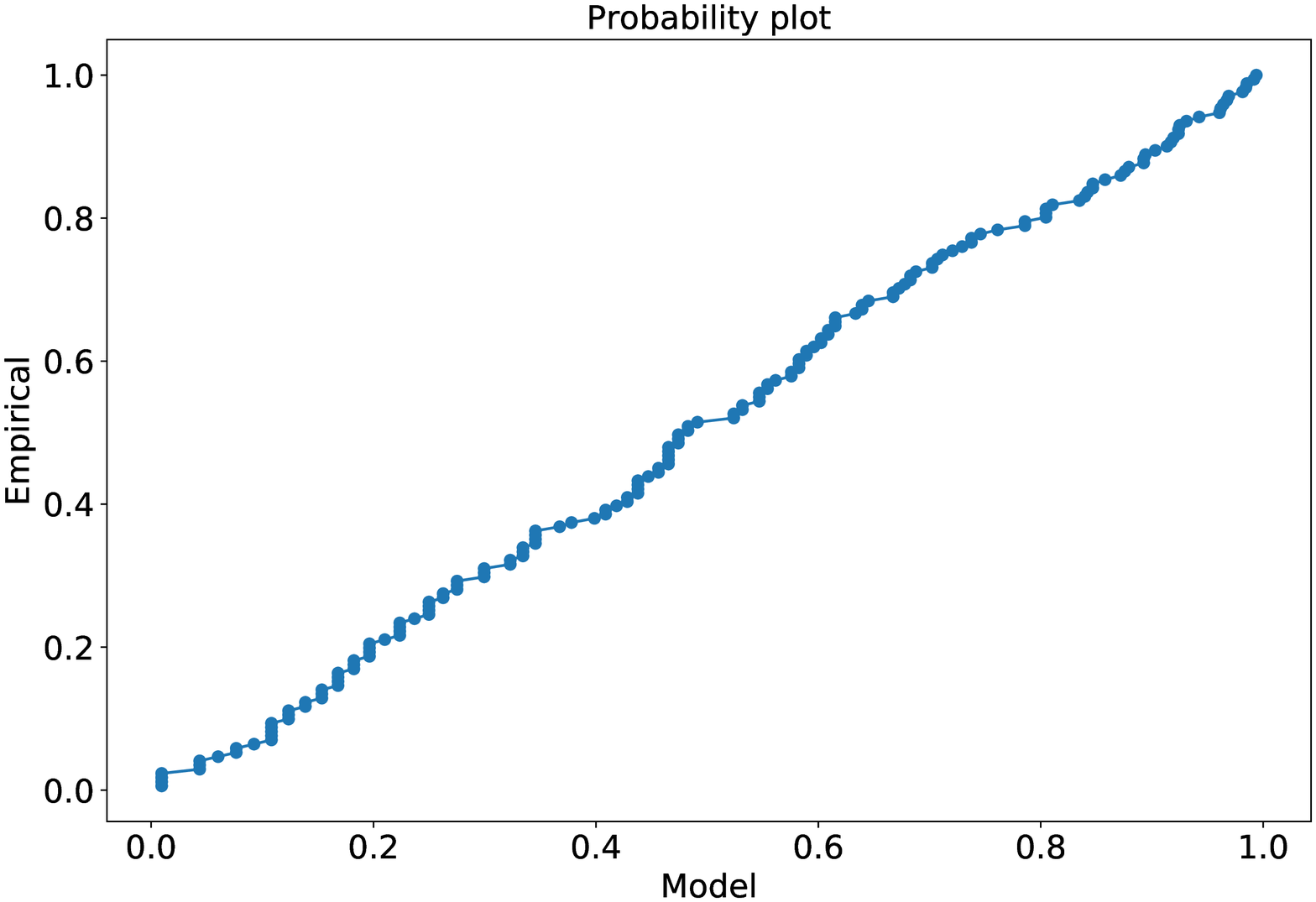}
        \end{center}
      \end{minipage}

      \begin{minipage}{0.5\hsize}
        \begin{center}
          \includegraphics[clip, width=6cm]{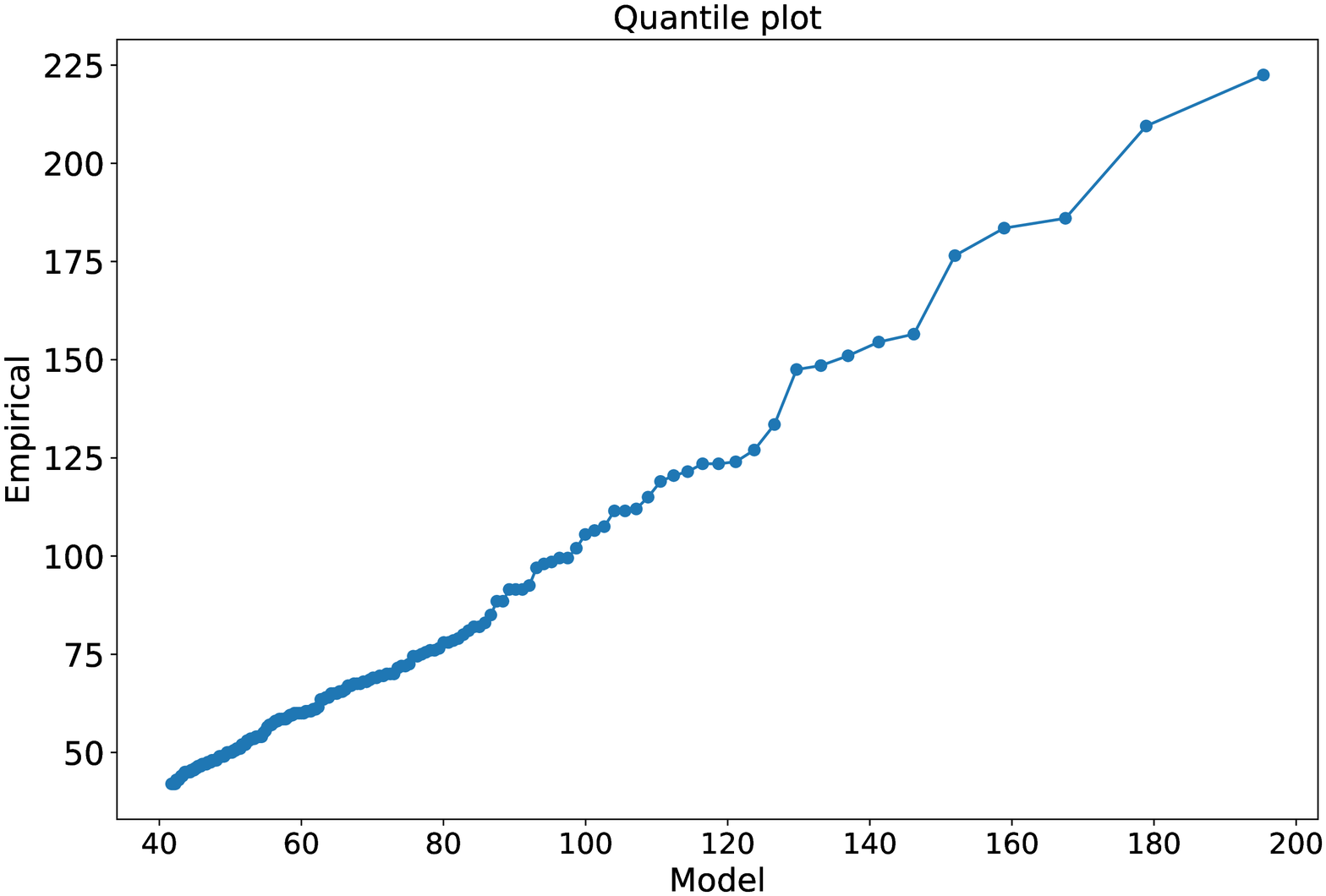}
        \end{center}
      \end{minipage}\\
      \begin{minipage}{0.5\hsize}
        \begin{center}
          \includegraphics[clip, width=6cm]{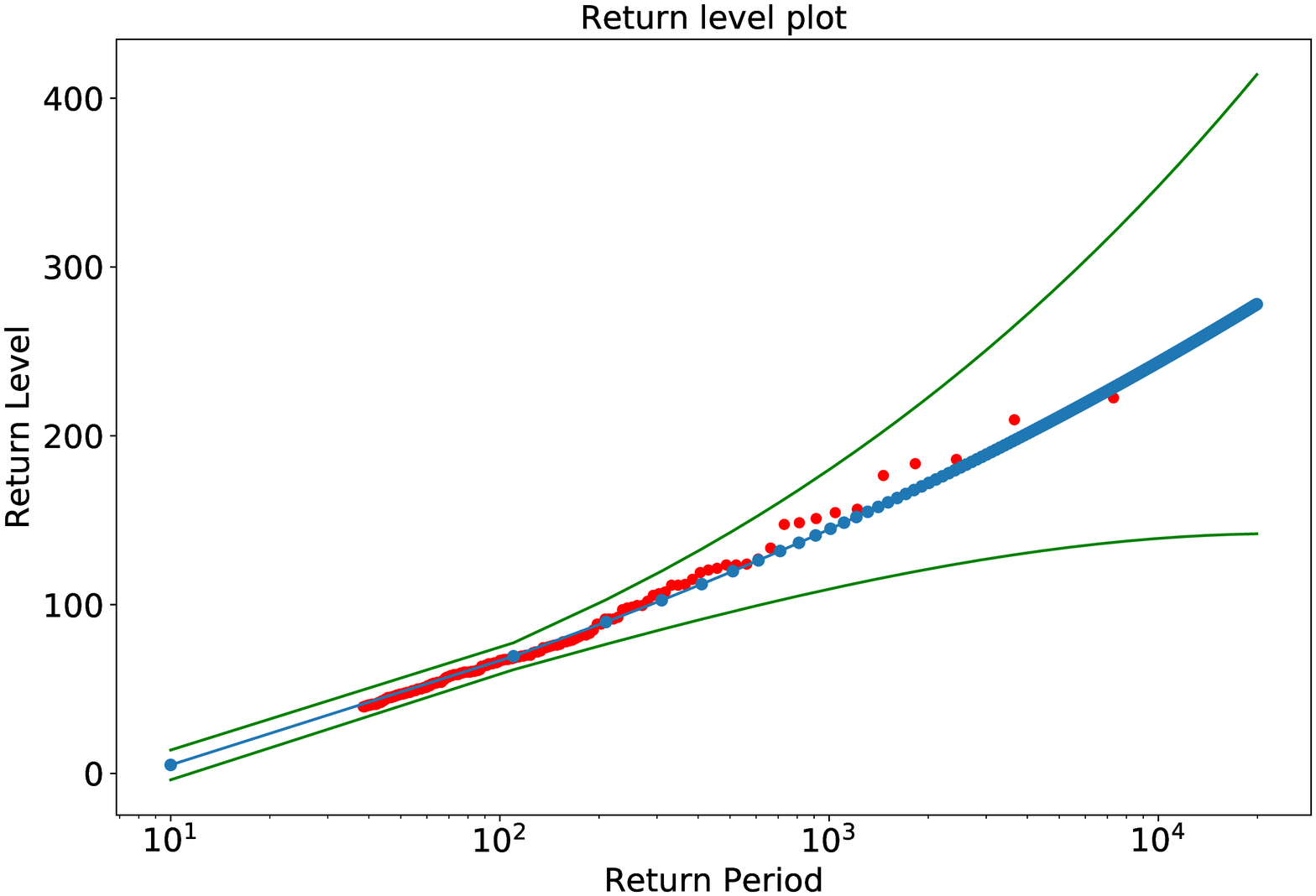}
        \end{center}
      \end{minipage}

      \begin{minipage}{0.5\hsize}
        \begin{center}
          \includegraphics[clip, width=6cm]{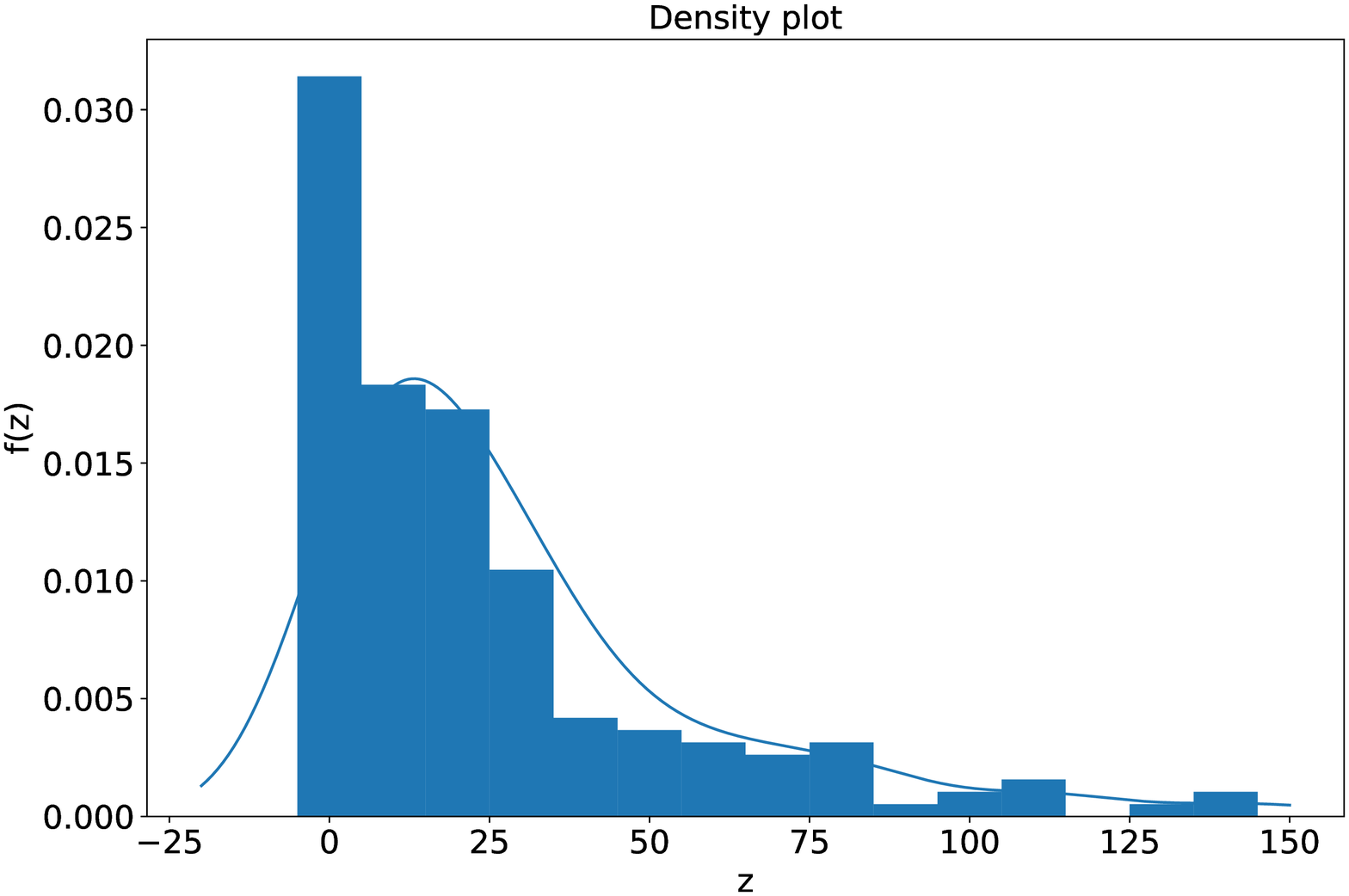}
        \end{center}
      \end{minipage}\\

    \end{tabular}
    \caption{Illustrations of probability plot, quantile plot, return level plot and density plot for rainfall data at Tokyo.}
    \label{Fig:Check_Tokyo}
  \end{center}
\end{figure}

We select the search range of threshold $u$ based on linearity of the mean excess plot and stability of $\widehat{\xi}$ and $\widehat{\sigma}^{\ast}$.
Then, we apply BO and search the value of threshold.
Figures  \ref{Fig:BO_Real} and  \ref{Fig:Check_Tokyo} illustrate the result of BO and fitness to GP model described in Section \ref{Sec:Check}.

The results of threshold, $\hat{\xi}$ and $\hat{\sigma}$ are summarized in Table \ref{Table:Real}.
\begin{table}[htbp]  \label{Table:Real}
\begin{center}
\caption{Results of BO for rain fall datas}
\begin{tabular}{|c|c|c|c|c|c|}
\hline Probability & domain (of BO) & Trial & threshold & $\hat{\xi}$ & $\widehat{\boldsymbol{\sigma}}$ \\
\hline Tokyo & (30,60) & 1 & 38.373 & 0.1153 & 26.5021 \\
\cline { 3 - 6 }      &         & 2 & 39.093 & 0.1032 & 27.1801 \\
\cline { 3 - 6 }      &         & 3 & 41.157 & 0.0676 & 29.2982 \\
\hline Osaka & (30,40) & 1 & 32.001 & 0.1812 & 17.2531 \\
\cline { 3 - 6 }& & 2 & 31.524 & 0.1785 & 17.2499 \\
\cline { 3 - 6 }& & 3 & 31.300 & 0.1993 & 16.5115 \\
\hline Fukuoka & (30,50) & 1 & 46.022 & 0.0154 & 31.1091 \\
\cline { 3 - 6 }& & 2 & 46.726 & 0.0257 & 30.5055 \\
\cline { 3 - 6 }& & 3 & 44.709 & 0.0125 & 31.2705 \\
\hline
\end{tabular}
\end{center}
\end{table}

\section{Discussion and future work} \label{Sec:Discuss}
This paper has proposed a machine-learning based method for  choice of hyper-parameters in EVT.
One of the key points of our approach is introducing an additional criterion and determining the value of hyper-parameter based on  machine-learning techniques.

\medskip
In EVT, there are other methods for the analysis of extreme events such as block maxima method,
  order statistics of extreme value theory and Poisson point process approach.
One of the challenging points is that we need to choose discrete hyper-parameters for these approaches
  whereas the hyper-parameter is continuous in POT method.
It is interesting to investigate methods of choice of discrete hyper-parameters.


\begin{thebibliography}{99}
\bibitem{BH74} A. A. Balkema and L. De Haan. "Residual life time at great age". The Annals of probability, (1974).



\bibitem{BYZ18} Bader, Brian, Jun Yan, and Xuebin Zhang. "Automated threshold selection for extreme value analysis via ordered goodness-of-fit tests with adjustment for false discovery rate." The Annals of Applied Statistics 12.1 (2018): 310-329.



\bibitem{CBTD01}  Coles, Stuart, et al. An introduction to statistical modeling of extreme values. Vol. 208. London: Springer, 2001.

\bibitem{CG16}
Frederico Caeiro and M. Ivette Gomes. "Threshold selection in extreme value analysis." Extreme value modeling and risk analysis: Methods and applications (2015): 69-82.


\bibitem{CLZP15}  Chen, Jiangpeng, et al. "Using extreme value theory approaches to forecast the probability of outbreak of highly pathogenic influenza in Zhejiang, China." PloS one 10.2 (2015): e0118521.

\bibitem{CS01} Vartan Choulakian and Michael A. Stephens. "Goodness-of-fit tests for the generalized Pareto distribution." Technometrics 43.4 (2001): 478-484.



\bibitem{DS90}
Davison, Anthony C., and Richard L. Smith. "Models for exceedances over high thresholds." Journal of the Royal Statistical Society: Series B (Methodological) 52.3 (1990): 393-425.


\bibitem{DuMouchel83} DuMouchel, William H. "Estimating the stable index $\alpha $ in order to measure tail thickness: A critique." the Annals of Statistics 11.4 (1983): 1019-1031.


\bibitem{EE20} Vignotto, Edoardo, and Sebastian Engelke. "Extreme value theory for anomaly detection–the GPD classifier." Extremes 23.4 (2020): 501-520.


\bibitem{EKM13} Embrechts, Paul, Claudia Klüppelberg, and Thomas Mikosch. Modelling extremal events: for insurance and finance. Vol. 33. Springer Science \& Business Media, 2013.

\bibitem{FDP03} Ferreira Ana, Laurens de Haan, and Liang Peng,  "On optimising the estimation of high quantiles of a probability distribution." Statistics 37.5 (2003): 401-434.

\bibitem{FT28} Fisher, Ronald Aylmer, and Leonard Henry Caleb Tippett. "Limiting forms of the frequency distribution of the largest or smallest member of a sample." Mathematical Proceedings of the Cambridge Philosophical Society. Vol. 24. No. 2. Cambridge University Press, 1928.

\bibitem{Gnedenko43} B. Gnedenko. Sur la distribution limite du terme maximum d’une serie aleatoire. Annals of
mathematics, pages 423–453, 1943.

\bibitem{Hill75} Bruce M. Hill.  "A simple general approach to inference about the tail of a distribution." The annals of statistics (1975): 1163-1174.


\bibitem{KZZH07}
Kharin, Viatcheslav V., et al. "Changes in temperature and precipitation extremes in the IPCC ensemble of global coupled model simulations." Journal of Climate 20.8 (2007): 1419-1444.

\bibitem{LMPD16} Langousis Andreas et al. "Threshold detection for the generalized Pareto distribution: Review of representative methods and application to the NOAA NCDC daily rainfall database." Water Resources Research 52.4 (2016): 2659-2681.


\bibitem{Marco14}
Rocco, Marco. "Extreme value theory in finance: A survey." Journal of Economic Surveys 28.1 (2014): 82-108.

\bibitem{Pickands75} J. Pickands III. Statistical inference using extreme
order statistics. the Annals of Statistics, 1975.


\bibitem{SFTL17} Siffer, Alban, et al. "Anomaly detection in streams with extreme value theory." Proceedings of the 23rd ACM SIGKDD International Conference on Knowledge Discovery and Data Mining. 2017.

\bibitem{SM12} Carl Scarrott and Anna MacDonald.
  "A review of extreme value threshold estimation and uncertainty quantification." REVSTAT–Statistical Journal 10.1 (2012): 33-60.


\bibitem{YXY18} Yuanyan, Luo, Du Xuehui, and Sun Yi. "Data streams anomaly detection algorithm based on self-set threshold." Proceedings of the 4th International Conference on Communication and Information Processing. 2018.


\end{thebibliography}
\end{document}